\documentclass{article}
\newif\ifanon
\ifdefined\FORNEURIPS
  \anontrue
\else
  \anonfalse
\fi
\ifanon
  \usepackage{neurips_2026}
\else
  \usepackage[preprint]{neurips_2026}
\fi
\usepackage[utf8]{inputenc}
\usepackage[T1]{fontenc}
\usepackage{amsmath,amssymb,amsfonts}
\usepackage{booktabs}
\usepackage{graphicx}
\usepackage{float}
\usepackage{xcolor}
\usepackage{hyperref}
\hypersetup{hidelinks}
\usepackage{multirow}

\usepackage{pgfplots}
\pgfplotsset{compat=1.18}
\usepgfplotslibrary{colorbrewer}
\usepackage{tikz}
\usetikzlibrary{arrows.meta,decorations.pathreplacing,calc}
\makeatletter
\@ifundefined{nolinenumbers}{}{}
\@ifundefined{linenumbers}{\newcommand{\linenumbers}{}}{}
\makeatother

\definecolor{heatblue}{RGB}{33,102,172}
\definecolor{heatlightblue}{RGB}{146,197,222}
\definecolor{heatneutral}{RGB}{247,247,247}
\definecolor{heatlightred}{RGB}{244,165,130}
\definecolor{heatred}{RGB}{202,0,32}
\definecolor{heatgray}{RGB}{220,220,220}

\definecolor{scaleorange}{RGB}{230,159,0}
\definecolor{scaleskyblue}{RGB}{86,180,233}
\definecolor{scalepurple}{RGB}{204,121,167}
\definecolor{scaleblue}{RGB}{0,114,178}

\title{When Does Removing LayerNorm Help?\\Activation Bounding as a Regime-Dependent Implicit Regularizer}

\ifanon
  \author{Anonymous Author(s) \\ Affiliation \\ \texttt{anonymous@domain.tld}}
\else
  \author{
    Lucky Verma \\
    Independent Researcher\\
    \texttt{luckyv1@umbc.edu}
  }
\fi

\begin{document}

\maketitle

\begin{abstract}
Dynamic Tanh (DyT) removes LayerNorm by bounding activations with a learned $\tanh(\alpha x)$. We show that this bounding is a regime-dependent implicit regularizer, not a uniformly beneficial replacement. Across GPT-2-family models spanning 64M--3.78B parameters and 1M--118M tokens, with Llama and ViT cross-checks, DyT improves validation loss by 27.3\% at 64M/1M but worsens it by 18.8\% at 64M/118M; the 1M benefit vanishes with capacity ($+$1.7\% at 3.78B), while the 118M penalty reaches $+$27.9\%. The mechanism is measurable: 49\% of DyT activations saturate at 1M versus 23\% at 118M, and a 500-step saturation heuristic classifies DyT's sign with 75\% raw in-sample accuracy on the 12-cell GPT-2 calibration set (AUC 0.75; 64\% when adding Scale~5 stress cells), correctly labels 3/3 Llama checks, but only reaches 50\% raw leave-one-scale-out accuracy. Three interventions support the bounding explanation: HardTanh reproduces the regime pattern, increasing $\alpha$ at 118M monotonically reduces DyT's penalty, and vanilla+dropout$(p{=}0.5)$ matches DyT's data-rich loss. We also localize Llama-DyT collapse to SwiGLU gating, where saturation separates collapse from convergence in a 3-seed component ablation ($r{=}0.94$). Scope: all experiments are compute-limited (T/P$<1.84$), below Chinchilla-optimal training.
\end{abstract}

\ifanon\else
\paragraph{Code and artifacts.}
Code, configurations, analysis scripts, and machine-readable result manifests are available at
\href{https://github.com/lucky-verma/dyt-composition-study}{\texttt{github.com/lucky-verma/dyt-composition-study}}.
\fi

\section{Introduction}

Normalization-free Transformers are now an active design space: DyT~\cite{zhu2025dyt}, Derf~\cite{derf2025normfree}, BHyT~\cite{bhyt2025}, TaperNorm~\cite{tapernorm2026}, and related placement variants ask \emph{what} replaces LayerNorm and \emph{where} it should sit. Existing evidence is mostly large-scale and single-regime; it rarely asks when removing normalization fails. This matters because the same bounded activation can be a useful capacity constraint in low-data training and an unnecessary bottleneck when data is sufficient.

We test this directly across GPT-2-family models, Llama-style cross-architecture validation, and a ViT/CIFAR-10 check. DyT's sign is regime-dependent: $-$27.3\% validation loss at 64M/1M, $+$18.8\% at 64M/118M, and $+$27.9\% at 3.78B/118M. The token-to-parameter ratio is already standard in scaling-law work~\cite{kaplan2020scaling,chinchilla2022}; our contribution is the crossover shape and its activation-saturation mechanism.

\begin{enumerate}
    \item \textbf{DyT is a regime-dependent implicit regularizer.} Its benefit attenuates with capacity and turns into a data-rich penalty; saturation falls from 49\% at 1M to 23\% at 118M, while HardTanh, $\alpha$-strength, and dropout-equivalence controls support activation bounding as the mechanism.

    \item \textbf{The effect transfers directionally but is architecture-sensitive.} GPT-2, ViT, RMSNorm, and Llama-style models show the same regime direction, but Llama-DyT has a $\sim$33\% per-seed collapse mode localized to SwiGLU gating; within this component ablation, saturation separates collapse from convergence with Pearson $r{=}0.94$.

    \item \textbf{A practitioner's screening recipe.} A 500-step saturation calibration gives 75\% raw in-sample accuracy on the 12-cell GPT-2 calibration set (AUC 0.75; 64\% with Scale~5 stress cells), correctly labels 3/3 Llama checks, but only 50\% raw leave-one-scale-out accuracy, so we present it as a scale-dependent screening heuristic rather than a universal rule.
\end{enumerate}

\medskip\noindent\textbf{Scope.} We study the compute-limited regime (T/P $<1.84$), not Chinchilla-optimal pretraining (T/P $\approx$20). The large-scale DyT parity results~\cite{zhu2025dyt} and our low-data failure modes are therefore complementary.

\section{Background and Related Work}

\paragraph{Normalization-free training.} DyT~\cite{zhu2025dyt} replaces LayerNorm with $\gamma\tanh(\alpha x)+\beta$ and reports large-scale parity on ViTs, LLaMA, wav2vec~2.0, and DiT. Derf~\cite{derf2025normfree}, BHyT~\cite{bhyt2025}, TaperNorm~\cite{tapernorm2026}, and related placement work~\cite{kim2025periln,karagodin2025normattndyn} expand this design space, while \citet{ziomek2025onelayernorm,singhal2025lnmem} connect normalization to memorization and generalization. Recent APJN theory predicts subcritical signal propagation for DyT/Derf-style replacements at initialization~\cite{alekseev2026subcritical}; we complement this by varying data scale and measuring trained-model saturation when bounding helps or hurts.

\paragraph{Attention modifications and systematic studies.} Differential Attention~\cite{ye2025diffattn} subtracts two softmax maps and improves large-scale long-context behavior; V2~\cite{diffattnv2}, DINT~\cite{dint2025}, and GDA~\cite{gda2025} refine the attention mechanism without testing data-regime dependence. \citet{narang2021transfer} and \citet{qiu2025gatedattn} provide systematic modification studies, but not across normalization and attention axes under matched data regimes.

\paragraph{Regime-dependent regularization.} Weight decay can operate through opposite mechanisms in over- and under-training~\cite{neurips2024weightdecay}; \citet{abouzeid2026optimizer} separately identify activation saturation as a candidate mediator under optimizer changes. Our data-budget sweep shows the same duality for a fixed optimizer and architectural intervention.

\section{Experimental Setup}
\label{sec:setup}

We study two modifications that target independent Transformer subsystems: normalization (DyT~\cite{zhu2025dyt} replacing LayerNorm) and attention (Differential Attention~\cite{ye2025diffattn}). Each is the most actively researched 2024--2025 proposal in its category, each is advertised as a drop-in replacement, and each has been evaluated primarily at a single data scale. We additionally include RMSNorm~\cite{zhang2019rmsnorm} as a second normalization baseline and a Vision Transformer cross-architecture probe.

\paragraph{Implementation.} All modifications are toggle flags in a single nanoGPT-based~\cite{karpathy2023nanogpt} \texttt{model.py} (following the nGPT~\cite{loshchilov2025ngpt} precedent), sharing identical optimizer, data pipeline, and training loop. DyT replaces LayerNorm with $\tanh(\alpha x) \cdot \gamma + \beta$ with learnable scalar $\alpha$. DiffAttn computes attention as the difference of two softmax maps with a learnable $\lambda$ and a GroupNorm stabilizer.

\paragraph{Configurations.} At each scale we train Vanilla (LayerNorm), DyT, and DiffAttn; RMSNorm is run at 64M only. Three seeds per condition. We further run an $\alpha$-initialization sweep (Appendix~\ref{app:alpha}), a 64M/118M dropout sweep, and a ViT-Small cross-validation.

\paragraph{Scales and data regimes.} Table~\ref{tab:scales} summarizes the five GPT-2-family model scales (64M--3.78B parameters) and four data budgets (1M--118M Wikitext-103 tokens) studied. Together these span a $58{\times}$ model-capacity range and a $118{\times}$ data range, covering the T/P ratio regime from deeply overparameterized (T/P$=2.6{\times}10^{-4}$ at 3.78B/1M) to moderately data-rich (T/P$=1.84$ at 64M/118M).

\begin{table}[ht]
\centering
\small
\caption{Experimental scales and data regimes. Parameter counts are measured totals for the nanoGPT family with tied embeddings; DiffAttn adds $\sim$10--20\% parameters via the second Q/K/V branch.}
\label{tab:scales}
\begin{tabular}{lrlll}
\toprule
\textbf{Scale} & \textbf{Params} & \textbf{Config ($L$/$H$/$D$)} & \textbf{Data (tokens)} & \textbf{Seeds} \\
\midrule
Scale~1 & 64M  & 12 / 8 / 512   & 1M / 10M / 50M / 118M & 3 \\
Scale~2 & 124M & 12 / 12 / 768  & 1M / 10M / 118M       & 3 \\
Scale~3 & 354M & 24 / 16 / 1024 & 1M / 10M / 118M       & 3 \\
Scale~4 & 1.3B & 24 / 32 / 2048 & 1M / 10M / 118M       & 3 \\
Scale~5$^\dagger$ & 3.78B & 32 / 32 / 3072 & 1M / 118M  & 3 \\
ViT     & 5.2M & 6 / 4 / 256    & CIFAR-10              & 1 \\
\bottomrule
\end{tabular}
\end{table}

\paragraph{Training details.} AdamW~\cite{loshchilov2019adamw} optimizer, learning rate $3 \times 10^{-4}$ (1e-4 for 1.3B and 3.78B) with cosine decay, sequence length 512, bfloat16 precision, \texttt{torch.compile} enabled. No dropout (except in dropout baseline comparison). Primary runs trained for 5,000 steps with evaluation every 500 steps. Batch sizes adjusted per scale: 64 (64M), 32 (124M), 16 (354M), 4 with gradient accumulation 16 (1.3B), 1 with gradient accumulation 64 (3.78B). Data: Wikitext-103 subsets (1M/10M/50M/118M tokens) prepared via BPE tokenization (GPT-2 vocabulary, 50,257 tokens).

\footnotetext[1]{$\dagger$~Scale~5 keeps the GPT-2 architecture family fixed while extending to 3.78B measured parameters; Llama-style cross-architecture validation is separate (Section~\ref{sec:cross}).}

\section{Results}
\label{sec:results}

Sections~\ref{sec:phase}--\ref{sec:llama_instability} make the empirical case: DyT's effect is regime-dependent, mechanistically tied to activation saturation, and architecture-sensitive under Llama-style SwiGLU gating. Section~\ref{sec:predictor_section} then turns these findings into a screening recipe and adds mechanistic probes with HardTanh, alpha-strength, and dropout-equivalence controls. Together, the evidence supports one argument: T/P helps organize when DyT's activation bounding helps or hurts, but the practical decision should be screened by short calibration before committing to full training.

\subsection{Phase Diagram: DyT Benefit Is Regime-Dependent}
\label{sec:phase}

\begin{figure}[t]
\centering
\includegraphics[width=0.98\textwidth]{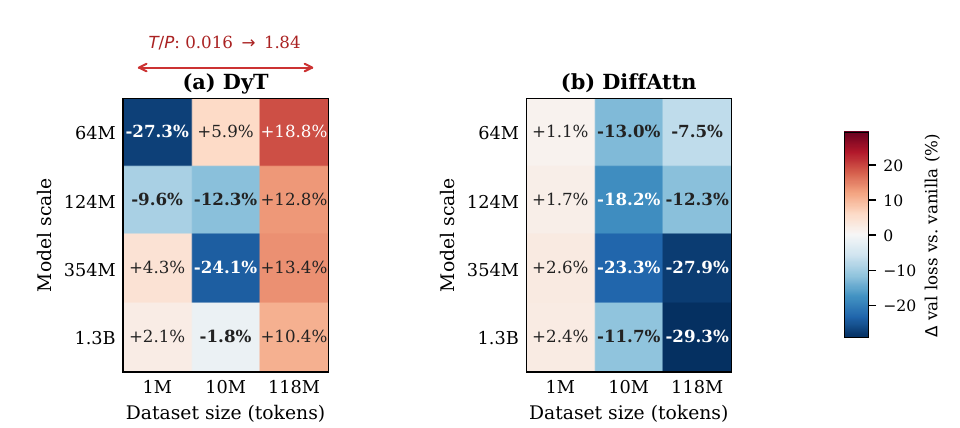}

\caption{\textbf{Phase diagrams reveal opposite regime preferences.} $\Delta$ validation loss vs.\ vanilla across the four primary GPT-2-family scales (64M--1.3B) and data regimes (3 seeds each); the Scale~5 stress test appears in Table~\ref{tab:scaling} and Appendix~\ref{app:scaling_curve}, Figure~\ref{fig:scaling}. Bold cells indicate DyT/DiffAttn helps; regular cells indicate it hurts. \textbf{(a)}~DyT acts as an implicit regularizer only when overparameterized (top-left, $-$27.3\%), with the effect vanishing as model capacity increases. At 354M/10M, DyT achieves its strongest benefit ($-$24.1\%), showing that the crossover shifts right with capacity. \textbf{(b)}~Differential Attention shows the opposite pattern: benefit grows with both data and model scale ($-$29.3\% at 1.3B/118M; all cells 3-seed means). The T/P ratio helps organize which modification helps.}
\label{fig:phase}
\end{figure}

We start with the central result (Table~\ref{tab:phase}, Figure~\ref{fig:phase}): a two-dimensional sweep of DyT and DiffAttn across dataset sizes, with a scaling study (Table~\ref{tab:scaling}) extending the sweep to larger models.

\begin{table}[ht]
\centering
\caption{Phase diagram: validation loss at 5K steps across four data scales (64M parameter model, 3 seeds). $\Delta$ columns show percentage change vs.\ vanilla.}
\label{tab:phase}
\begin{tabular}{lrrrrrrr}
\toprule
\textbf{Data} & \textbf{Vanilla} & \textbf{DyT} & \textbf{DiffAttn} & \textbf{$\Delta$ DyT} & \textbf{$\Delta$ DiffAttn} & \textbf{Regime} \\
\midrule
1M   & 9.384{\scriptsize$\pm$.04} & \textbf{6.819}{\scriptsize$\pm$.17} & 9.490{\scriptsize$\pm$.10}  & $-$27.3\% & $+$1.1\%  & Overfit \\
10M  & 4.260{\scriptsize$\pm$.01} & 4.510{\scriptsize$\pm$.01}          & \textbf{3.706}{\scriptsize$\pm$.03} & $+$5.9\%  & $-$13.0\% & Moderate \\
50M  & 3.666{\scriptsize$\pm$.01} & 4.386{\scriptsize$\pm$.02}          & \textbf{3.380}{\scriptsize$\pm$.01} & $+$19.7\% & $-$7.8\%  & Mild underfit \\
118M & 3.631{\scriptsize$\pm$.01} & 4.313{\scriptsize$\pm$.02}          & \textbf{3.359}{\scriptsize$\pm$.01} & $+$18.8\% & $-$7.5\%  & Underfit \\
\bottomrule
\end{tabular}
\end{table}

\begin{table}[ht]
\centering
\small
\setlength{\tabcolsep}{4pt}
\caption{DyT and DiffAttn effect across model scales (3 seeds, eff\_batch$=$64). DyT's 1M benefit attenuates with capacity; DiffAttn helps in data-rich cells through 1.3B. At Scale~5, the 118M DiffAttn V1 and sigmoid-$\lambda$ ablation cells enter high-loss collapse while the 1M cells are harmful but finite; the 3.78B/10M cell was not run because Scale~5 is a 1M/118M stress test (Appendix~\ref{app:sigmoid_lambda_diffattn}). $^\S$S5 DyT raw $p{=}0.004$, Bonferroni $p{=}0.078$.}
\label{tab:scaling}
\begin{tabular}{lrrrr}
\toprule
\textbf{Scale} & \textbf{DyT 1M} & \textbf{DyT 10M} & \textbf{DyT 118M} & \textbf{DiffAttn 118M} \\
\midrule
64M   & $-$27.3\%  & $+$5.9\%  & $+$18.8\% & $-$7.5\%  \\
124M  & $-$9.6\%   & $-$12.3\% & $+$12.8\% & $-$12.3\% \\
354M  & $+$4.3\%   & $-$24.1\% & $+$13.4\% & $-$27.9\% \\
1.3B  & $+$2.1\%   & $-$1.8\%  & $+$10.4\% & $-$29.3\% \\
3.78B$^\dagger$ & $+$1.7\%   & \textit{not run} & $+$27.9\%\,$^\S$ & \textit{high-loss collapse}\,$^\ddagger$ ($+$207\%) \\
\bottomrule
\end{tabular}
\footnotetext[2]{$\ddagger$~Scale~5 DiffAttn V1 and sigmoid-$\lambda$ ablation 118M losses are 10.54{\scriptsize$\pm$1.45} and 12.72{\scriptsize$\pm$4.13} vs.\ vanilla 3.43. Collapse denotes finite but severe high-loss optimization failure after completed 5K-step runs, not OOM/job failure; full per-seed analysis appears in Appendix~\ref{app:sigmoid_lambda_diffattn}.}
\end{table}

\paragraph{Statistical significance.} Across 19 paired vanilla-vs-modification comparisons, 13 survive Bonferroni correction at $p_{\text{Bonferroni}}{<}0.05$ (Appendix~\ref{app:sig_tests}). The core DyT sign flip is significant at Scale~1: $-27.3\%$ at 1M ($p_{\text{Bonf}}{=}0.032$) and $+18.8\%$ at 118M ($p_{\text{Bonf}}{=}0.020$). The S5/118M DyT penalty is directionally large ($+27.9\%$; raw $p{=}0.004$) but marginal after correction ($p_{\text{Bonf}}{=}0.078$), so we treat it as scale-consistent stress-test evidence rather than a standalone claim.

Three findings emerge from Tables~\ref{tab:phase}--\ref{tab:scaling} and Figure~\ref{fig:phase} (model-scale visualization in Appendix~\ref{app:scaling_curve}, Figure~\ref{fig:scaling}).

\textbf{1. DyT crosses from beneficial to harmful between 1M and 10M tokens at Scale~1.} At 64M/1M, vanilla memorizes (train loss 0.12, train/val gap 9.22) while DyT holds train loss at 2.47 (gap 4.31), yielding 27.3\% lower validation loss. At 10M tokens and beyond, overfitting is no longer the bottleneck and DyT's convergence penalty dominates.

\textbf{2. The effect depends on both data and capacity.} Holding tokens at 118M, DyT is harmful at every scale; holding tokens at 1M, its benefit attenuates from $-$27.3\% at 64M to neutral/harmful at 354M--3.78B. The strongest benefit appears at 354M/10M ($-$24.1\%), showing that the crossover shifts with capacity rather than following a fixed token count.

\textbf{3. Scale~5 is a stress test, not a new calibration range.} At 3.78B, DyT is neutral at 1M ($+$1.7\%) and harmful at 118M ($+$27.9\%, $p_{\text{Bonf}}{=}0.078$), consistent with the lower-saturation forecast. DiffAttn V1 and a V2-inspired sigmoid-$\lambda$ ablation both complete to 5K; the 1M cells are harmful ($+$34.8\% V1, $+$25.2\% ablation), while the 118M cells enter high-loss collapse ($+$207\% V1, $+$271\% ablation; Appendix~\ref{app:sigmoid_lambda_diffattn}). We therefore do not extrapolate DiffAttn scaling beyond 1.3B under this 5K-step budget.

\textbf{DiffAttn shows the opposite regime preference.} Its dual-softmax noise cancellation is irrelevant at 1M tokens ($+$1.1\%) but provides substantial gains at 10--118M tokens ($-$13.0\% at 10M, $-$7.8\% at 50M, $-$7.5\% at 118M). The benefit \emph{grows with model capacity}: $-$7.5\% at 64M, $-$12.3\% at 124M, $-$27.9\% at 354M, $-$29.3\% at 1.3B (Table~\ref{tab:scaling}), mirroring DyT's weakening. Where DyT is a capacity-\emph{constraining} modification, DiffAttn is a capacity-\emph{enhancing} one.

\textbf{The two modifications solve different problems.} At 1M tokens the bottleneck is overfitting, and only DyT helps; at 10M tokens and beyond the bottleneck is model quality, and only DiffAttn helps. The regime should therefore condition architectural choice.

\textbf{The crossover shifts right with model capacity.} At 354M/10M (T/P $=$ 0.03), DyT achieves its strongest benefit in the study, $-$24.1\%. This is counterintuitive from 64M results alone, where DyT hurts at 10M tokens ($+$5.9\%). Our interpretation is that 10M tokens is $35\times$ more overparameterized at 354M than at 64M (6.4$\times$); the tanh bottleneck remains active. The crossover is therefore not a fixed ratio; it shifts with capacity, a relationship we operationalize as the saturation heuristic in Section~\ref{sec:predictor}.

\paragraph{Extended training and composition dynamics.} A 10K-step extension at 64M/118M shows that DyT+DiffAttn's 5K-budget deficit is a convergence-horizon effect: it moves from 4.246{\scriptsize$\pm$.135} best validation loss under the 5K budget to 3.384{\scriptsize$\pm$.022} under the 10K budget, matching vanilla 3.388, while DiffAttn alone continues to improve to 2.926{\scriptsize$\pm$.02}. Full 3-seed table and representative curves are in Appendix~\ref{app:convergence}.

\paragraph{DyT initialization sensitivity.}
The $\alpha$ parameter controls regularization strength monotonically (lower $\alpha$ = stronger saturation = more regularization), with $\alpha\in[0.5,1.0]$ giving the strongest 64M/1M point estimates in our 2-seed sweep ($\approx -34\%$). The relationship extends to vision: ViT with $\alpha$=0.5 achieves $+$8.8\% over LayerNorm on CIFAR-10 (Appendix~\ref{app:alpha}).

\subsection{Mechanistic Verification: Why Does the Regime Shift Exist?}
\label{sec:mechanism}

The phase diagram establishes \emph{that} DyT's effect is regime-dependent. This section establishes \emph{why}: tanh saturation creates a data-dependent capacity bottleneck, and this bottleneck is directly measurable. We use two independent instruments, activation saturation statistics and weight spectral analysis, which both point to the same mechanistic story.

Does the regime dependence reflect implicit regularization, or is it merely slower convergence? We distinguish the two by measuring the fraction of tanh activations that are saturated ($|\alpha x| > 2.0$) across checkpoints (Table~\ref{tab:saturation_full}). At 1M tokens, roughly half of all DyT activations saturate: the tanh function clips them to $\pm$1 regardless of input magnitude, reducing the model's representational capacity. At 118M tokens, saturation drops to 23\%: the tanh operates in its near-linear regime and DyT behaves like LayerNorm. The capacity bottleneck is therefore data-dependent, consistent with implicit regularization and inconsistent with a pure convergence-rate explanation.

The train/val gap supports the mechanism directly. At 1M tokens, vanilla memorizes the training set (train loss $=$ 0.12, train/val gap $=$ 9.22) while DyT prevents memorization (train loss $=$ 2.47, gap $=$ 4.31), yielding 27.3\% better validation loss (Appendix~\ref{app:trainval}).

\paragraph{Independent verification: HTSR weight spectral analysis.} Heavy-Tailed Self-Regularization (HTSR)~\cite{martin2021predicting,weightwatcher2023} agrees with the activation-saturation picture across 9 of 10 (scale, tokens) cells: DyT produces lower power-law exponent $\bar{\alpha}$ (more structured weights) than vanilla. The Scale~5/118M cell illustrates the limit of HTSR as a generalization proxy when bounded activations \emph{impose} structure rather than letting it emerge: DyT's $\bar{\alpha}=4.04$ (92\% layers in power-law regime) vs.\ vanilla $\bar{\alpha}=5.44$ (63\%), yet DyT validation loss is 27.9\% \emph{worse}: structure without learning the loss landscape. Full per-layer $\alpha$ + reversal at Scale~5/1M: Appendix~\ref{app:weightwatcher}.

\subsection{Cross-Architecture and Cross-Normalization Validation}
\label{sec:cross}

Sections~\ref{sec:phase}--\ref{sec:mechanism} establish the phase diagram and its mechanism on GPT-2. Is the regime dependence an artifact of that particular architecture, or a general property of bounded activations? We test generalization along two axes: \emph{architecture} (Llama-style RoPE+SwiGLU+GQA and a Vision Transformer) and \emph{normalization} (RMSNorm as a second baseline).

\paragraph{Vision Transformer.} We train ViT-Small~\cite{dosovitskiy2021vit} (6 layers, 256-dim) on CIFAR-10. CIFAR-10 at this scale is an overfitting regime (50K images, easily memorized), the vision analogue of our 1M-token language setting. DyT at $\alpha$=0.5 reaches 81.2\% validation accuracy versus 72.4\% for LayerNorm, a $+$8.8-point gain directionally consistent with the language-modeling result (full $\alpha$ sweep in Appendix~\ref{app:alpha}).

\paragraph{RMSNorm baseline.} RMSNorm matches LayerNorm within 0.7\% across all cells (Appendix~\ref{app:rmsnorm}); the DyT effect is the same whether the baseline normalizer is LayerNorm or RMSNorm.

\paragraph{Llama-style architecture (RoPE + SwiGLU + GQA).} On Llama~\cite{touvron2023llama,su2021rope,shazeer2020swiglu,ainslie2023gqa} (3 seeds), the regime pattern replicates: DyT improves val 25.6\% at 64M/1M and degrades 59.1\% at 64M/118M (one of three seeds unstable, see Section~\ref{sec:llama_instability}). DiffAttn mirrors this pattern: neutral at 1M ($+$0.9\%) and beneficial at 118M ($-$7.7\%). At Scale~2 (124M), DyT's 1M benefit attenuates to $-$7.1\% (GPT-2: $-$9.6\%); per-cell values + 3-seed std in Appendix~\ref{app:llama}, Table~\ref{tab:llama}.

\subsection{DyT Instability on Modern Architectures}
\label{sec:llama_instability}

\paragraph{DyT instability on Llama (33\% per-seed failure rate).} Across 6 DyT runs (3 seeds $\times$ 2 data regimes at Scale~1, $\sim$89M Llama-family params), 2 fail to converge, a 33\% catastrophic failure rate, versus no plateau-at-initialization failures observed in our GPT-2-family DyT runs at the same 5K-step budget. Failed runs plateau at initialization (train loss $\approx$7.4, val matches train, no gradient explosion); std at Llama 64M/118M is $\pm$1.3 vs.\ vanilla's $\pm$.003, making the result highly seed-sensitive.

\paragraph{Architecture-specific ablation: SwiGLU is the destructive interaction.} 3-seed ablations at 64M/118M test which Llama component drives the collapse: \texttt{ablate\_swiglu} (GELU-gated FFN) is the only ablation where all 3 seeds converge uniformly (val 4.476{\scriptsize$\pm$.007}, $\sigma_{|\alpha x|>2}{=}0.257$); \texttt{ablate\_rope} and \texttt{ablate\_gqa} retain bimodal collapse (2/3 seeds fail). Per-seed saturation $\sigma$ separates collapse from convergence in this ablation (Pearson $r{=}0.94$, $n{=}9$); threshold $\sigma{\geq}0.5$ classifies these runs without error (4/4 hits, 0/5 false positives). A plausible mechanism is that SwiGLU's multiplicative gating $(xW_1) \odot \mathrm{SiLU}(xW_2)$ amplifies pre-activation magnitudes, pushing DyT into its flat tanh tail; GELU-gated FFN lacks this multiplicative interaction (full mechanism + per-seed values: Appendix~\ref{app:r5_llama}). The 500-step calibration thus serves as a saturation measurement, stability check, and architecture-specific collapse warning in the screening recipe below.

\paragraph{Cross-dataset validation: OpenWebText.} Replicating the phase diagram on OpenWebText (3-seed at 64M/1M and 64M/118M; full table Appendix~\ref{app:owt}), DyT's pattern holds: $-$31.7\% at 1M (Wikitext: $-$27.3\%) and $+$14.6\% at 118M (Wikitext: $+$18.8\%); DiffAttn replicates likewise. A stricter cross-domain forward-pass test (Wikitext-trained S1 evaluated on OWT validation text without retraining) preserves regime direction: DyT $-$19.5\% at 1M, $+$7.9\% at 118M. This indicates that regime dependence is a property of the learned representations, not OWT-specific training dynamics.

\section{A Practitioner's Framework: When to Remove Normalization}
\label{sec:predictor_section}

Building on the preceding results, this section first gives a practitioner-facing recipe for screening DyT before committing to a full training run, then tests the proposed capacity-bottleneck mechanism through HardTanh, alpha-strength, and dropout-equivalence controls. The recipe uses a 500-step calibration when available and a low-confidence T/P prior otherwise.

\paragraph{DyT screening recipe.}
This is a calibration heuristic, not an algorithmic contribution. It applies to pre-norm Transformer decoders trained with the optimizer and schedule in Section~\ref{sec:setup}; other architecture families should be treated as out-of-distribution until calibrated. The ViT/CIFAR-10 result is a directional cross-check, not a validated recipe domain. The calibration cost is 500 optimizer steps (10\% of our primary 5K-step budget), plus saturation measurement on sampled text batches.
\begin{enumerate}
    \item Run a 500-step DyT calibration on the target model and data. Use at least two seeds for non-GPT-2-style stacks, and three when Llama-style RoPE+SwiGLU+GQA components are present.
    \item Measure saturation on sampled text batches:
    \[
    \sigma(M,X)=
    \frac{\sum_{\ell=1}^{L} |\{a \in A_\ell(X): |\alpha_\ell a|>2\}|}
         {\sum_{\ell=1}^{L}|A_\ell(X)|},
    \]
    where $A_\ell(X)$ are the DyT inputs at layer $\ell$ and $\alpha_\ell$ is that layer's learned DyT scale. This is the implemented global fraction of activations in the flat tanh tail; because all measured DyT layers share the same activation shape, it is numerically equivalent to averaging per-layer fractions. Our reported values use 50 sampled Wikitext forward passes at sequence length 512.
    \item If any calibration run plateaus near initialization, diverges, shows large seed-to-seed dispersion, or (for Llama-style stacks) reaches $\sigma{\geq}0.5$ in any seed, prefer LayerNorm/RMSNorm or add seeds before continuing DyT.
    \item Otherwise, if mean $\sigma>0.43$, DyT is a candidate worth continuing with validation monitoring; if mean $\sigma\leq0.43$, prefer LayerNorm/RMSNorm.
    \item Without calibration, use T/P only as a weak prior for GPT-2-style models below 354M parameters: T/P$<0.05$ favors trying DyT, T/P$>0.5$ favors LayerNorm/RMSNorm, and the middle region requires calibration. Do not use T/P-only selection for non-GPT-2 architectures or for $P\geq354$M.
\end{enumerate}

\paragraph{Validation summary.} The 0.43 saturation threshold classifies 9/12 pre-Scale-5 GPT-2 calibration cells (75\% raw, 68.8\% balanced, Wilson 95\% CI $[47\%, 91\%]$; AUC 0.75); including the two Scale~5 stress cells lowers this to 9/14 (64\% raw, AUC 0.60). LOSO cross-validation drops to 50\% raw (43.8\% balanced; Wilson $[25\%, 75\%]$), so the threshold value is scale-dependent and we treat the rule as a calibration heuristic, not a scale-invariant decision rule. The directional signal matches held-out Llama checks (3/3 correctly classified). Full numbers, per-fold thresholds, and residuals appear in Section~\ref{sec:predictor} and Appendix~\ref{app:predictor_detail}; the recipe above is the practitioner-facing summary.

\subsection{Saturation-Based Crossover Heuristic}
\label{sec:predictor}

\begin{figure}[H]
\centering
\begin{tikzpicture}
\begin{axis}[
    width=0.94\columnwidth,
    height=5.3cm,
    xlabel={Saturation fraction ($|\alpha x| > 2.0$)},
    ylabel={$\Delta$ validation loss vs.\ vanilla (\%)},
    xmin=0.15, xmax=0.55,
    ymin=-35, ymax=25,
    grid=both,
    grid style={gray!20},
    major grid style={gray!40},
    legend style={font=\scriptsize, fill=white, fill opacity=0.96, draw=gray!50,
                  rounded corners=2pt, cells={anchor=west}, legend columns=3,
                  column sep=0.7em, at={(0.5,1.04)}, anchor=south},
    every axis plot/.append style={only marks, mark size=3.5pt},
    extra y ticks={0},
    extra y tick style={grid=major, grid style={black, thick, dashed}},
    clip=true,  
]

\fill[heatred!6] (axis cs:0.15,-35) rectangle (axis cs:0.43,25);
\fill[heatblue!6] (axis cs:0.43,-35) rectangle (axis cs:0.55,25);
\fill[gray!20, opacity=0.32] (axis cs:0.27,-35) rectangle (axis cs:0.49,25);

\addplot[color=scaleorange, mark=*, mark options={fill=scaleorange}] coordinates {
    (0.493, -27.3) (0.413, 5.9) (0.237, 19.7) (0.234, 18.8)
};
\addlegendentry{GPT-2 64M}

\addplot[color=scaleskyblue, mark=square*, mark options={fill=scaleskyblue}] coordinates {
    (0.466, -9.6) (0.292, -12.3) (0.193, 12.8)
};
\addlegendentry{GPT-2 124M}

\addplot[color=scalepurple, mark=diamond*, mark options={fill=scalepurple}] coordinates {
    (0.490, 4.3) (0.369, -24.1) (0.327, 13.4)
};
\addlegendentry{GPT-2 354M}

\addplot[color=scaleblue, mark=pentagon*, mark options={fill=scaleblue}] coordinates {
    (0.393, 2.1) (0.238, 10.4)
};
\addlegendentry{GPT-2 1.3B}

\addplot[color=black, mark=triangle*, mark size=4pt, mark options={fill=none, thick}] coordinates {
    (0.536, -25.6) (0.452, -7.6)
};
\addlegendentry{Llama (cross-val)}
\addlegendimage{only marks, mark=o, mark size=4pt, black, thick}
\addlegendentry{Misclassified ($*$)}

\draw[->, black, thick] (axis cs:0.326, 16) -- (axis cs:0.326, 21);
\node[font=\scriptsize, black, anchor=west] at (axis cs:0.335, 21) {Llama +59.1\% off-scale};

\draw[ultra thick, black] (axis cs:0.292,-12.3) circle[radius=4.8pt];
\node[font=\bfseries\scriptsize, anchor=south west] at (axis cs:0.292,-12.3) {$\ast$};
\draw[ultra thick, black] (axis cs:0.490,4.3) circle[radius=4.8pt];
\node[font=\bfseries\scriptsize, anchor=south west] at (axis cs:0.490,4.3) {$\ast$};
\draw[ultra thick, black] (axis cs:0.369,-24.1) circle[radius=4.8pt];
\node[font=\bfseries\scriptsize, anchor=south west] at (axis cs:0.369,-24.1) {$\ast$};

\draw[thick, dashed, gray] (axis cs:0.43, -35) -- (axis cs:0.43, 25);
\node[font=\scriptsize, gray, anchor=south, rotate=90] at (axis cs:0.44, -5) {threshold = 0.43};

\node[font=\scriptsize, heatblue!70!black] at (axis cs:0.50, -32) {\textbf{DyT helps}};
\node[font=\scriptsize, heatred!70!black, anchor=west] at (axis cs:0.17, 22) {\textbf{DyT hurts}};

\end{axis}
\end{tikzpicture}
\caption{Activation saturation tracks DyT's effect. Light red/blue regions show the threshold rule's predicted harmful/beneficial regimes around the in-sample cutoff at 0.43; the gray band marks LOSO per-fold optima (0.27--0.49). Marker color/shape encodes model family and scale; black rings with $^{*}$ mark cells misclassified by the threshold rule. The threshold correctly classifies 9 of 12 pre-Scale-5 GPT-2 calibration cells (75\% raw, 68.8\% balanced; Wilson 95\% CI $[47\%, 91\%]$; 9/14 when Scale~5 stress cells are included) and all three Llama cross-validation points (triangles). The Llama 64M/118M point is annotated off-scale to preserve resolution near the GPT-2 decision boundary. Misclassified cells (S2/10M and two 354M cells) reflect the scale-dependent crossover shift.}
\label{fig:saturation_predictor}
\end{figure}
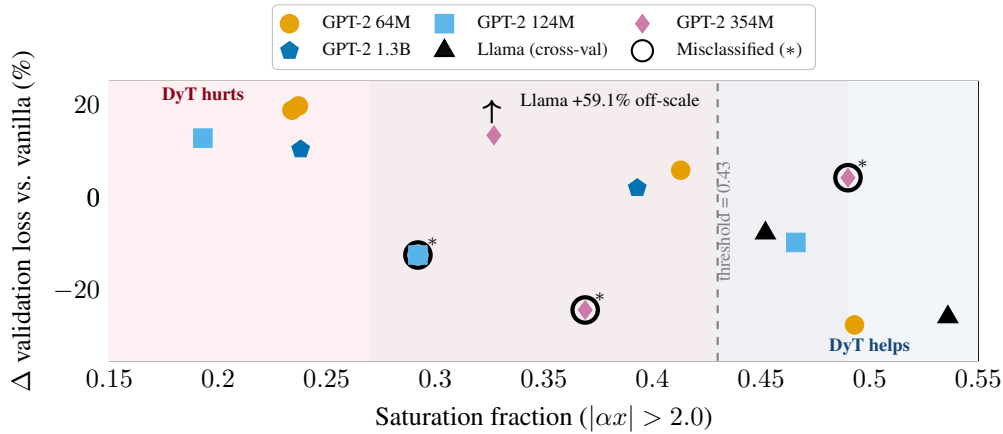

We measure activation saturation, the fraction of DyT activations with $|\alpha x| > 2.0$ that places them in the flat tanh tails, across 81 DyT checkpoints spanning five model scales, four data regimes, and two architectures (GPT-2 and Llama). Figure~\ref{fig:saturation_predictor} plots the relationship between saturation and DyT's validation effect; per-cell saturation values are listed in Appendix~\ref{app:predictor_detail}, Table~\ref{tab:saturation_full}.

\paragraph{Saturation $>$ 0.43 classifies 9 of 12 GPT-2 calibration cells.} The threshold rule \emph{DyT helps when sat $>$ 0.43} achieves 75\% raw in-sample accuracy on the pre-Scale-5 GPT-2 calibration set (68.8\% balanced, Wilson 95\% CI $[47\%, 91\%]$, AUC 0.75); including the two Scale~5 stress cells lowers the in-sample score to 64\% raw (9/14, AUC 0.60). Under leave-one-scale-out CV across S1--S4, pooled held-out accuracy drops to 50\% raw and 43.8\% balanced (Wilson $[25\%, 75\%]$); the threshold \emph{value} does not transfer cleanly across scales (per-fold optima 0.27--0.49). The directional signal is useful as a calibration cue, but the cutoff is scale-dependent. We therefore reframe the rule as a \textbf{calibration heuristic} for screening rather than a scale-invariant decision rule. The three GPT-2 misclassifications cluster at Scale~3 where 35$\times$-overparameterized cells show DyT benefit even at sat $<$ 0.43 (full per-cell residuals + a two-variable $(\text{sat}, \log P)$ linear fit yielding in-sample $R^2{=}0.42$ but Llama-OOD $R^2{=}-0.17$ in Appendix~\ref{app:predictor_detail}); this confirms the heuristic is a \emph{directional decision rule}, not a calibrated regression.

\paragraph{Out-of-distribution check.} The 0.43 threshold, calibrated entirely on GPT-2, classifies all three Llama cells correctly (sat $=$ 0.536/0.452/0.326 at 64M/1M, 124M/1M, 64M/118M); $n{=}3$ is narrow and sits clearly away from the decision boundary, so this is directional support rather than rigorous transfer. An interior cross-scale held-out point at Scale~2.5 (162.6M, not used in calibration) lands where the saturation trend suggests: $\Delta_{\text{DyT}}{=}-0.4\%$ at 1M (neutral, within seed noise) and $+$11.8\% at 118M (Appendix~\ref{app:scale25}). The 500-step calibration is the primary deployment tool; T/P is only a prior on which side of the threshold to expect. The $\sim$33\% Llama training instability (Section~\ref{sec:llama_instability}) is orthogonal to threshold accuracy and must be weighed separately.

\subsection{Intervention Evidence: Activation Bounding + Dropout-Equivalence}
\label{sec:causal}
\label{sec:hardtanh}
\label{sec:dropout_sweep}

Three interventions test whether \emph{activation bounding}, rather than the smooth $\tanh$ curve or the architectural identity of DyT, is the mechanism behind the regime-dependent effect.

\paragraph{Intervention 1: HardTanh (function-class control).} We replace $\tanh(\alpha x)$ with $\text{hardtanh}(x)$, a hard clip to $[-1,1]$. HardTanh reproduces the regime pattern directionally with a sharper bound: $-$33.4\% at 1M (DyT $-$27.3\%), $+$9.4\% at 10M (DyT $+$5.9\%), $+$25.6\% at 118M (DyT $+$18.8\%). The shared sign flip between DyT and HardTanh indicates the effect is a property of bounding, not the smooth $\tanh$ curve.

\paragraph{Intervention 2: $\alpha$ strength gives a data-rich dose response.} At 64M/118M, making DyT less bounded by increasing $\alpha$ monotonically reduces the penalty: $\alpha{=}0.5$ gives 5.154{\scriptsize$\pm$.045} ($+$42.0\%), $\alpha{=}1.0$ gives 4.771{\scriptsize$\pm$.046} ($+$31.4\%), $\alpha{=}2.0$ gives 4.313{\scriptsize$\pm$.024} ($+$18.8\%), and $\alpha{=}3.0$ gives 4.196{\scriptsize$\pm$.004} ($+$15.5\%). The best data-rich $\alpha$ still trails LayerNorm, so this is mechanistic dose-response evidence rather than a tuning win (Appendix~\ref{app:alpha}).

\paragraph{Intervention 3: Vanilla LayerNorm + dropout matches DyT at $p{\approx}0.5$.} If DyT's 118M penalty is regularization-type, an explicit regularizer at matched strength should reproduce it. Sweeping dropout $p \in \{0.1, 0.3, 0.5\}$ on vanilla LayerNorm at 64M/118M (3 seeds, eff\_batch=64; full table Appendix~\ref{app:dropout}), $p{=}0.5$ yields best val 4.295{\scriptsize$\pm$.007} during the 5K-step run, statistically indistinguishable from DyT's 4.313{\scriptsize$\pm$.02}. At lower $p$, dropout degrades vanilla by $+$1.2\%/$+$9.5\%; DyT skips these intermediate regimes and lands near the $p{=}0.5$ outcome. DyT therefore behaves like a moderately-high regularization setting in the data-rich regime where neither helps. What DyT contributes is not a \emph{better} regularizer but a \emph{simpler} one: the saturation fraction \emph{emerges} from the data ($49\%$ at 1M $\to$ $23\%$ at 118M; Section~\ref{sec:mechanism}) without a user-specified schedule.

\paragraph{Practical use.} Run DyT for 500 warm-up steps and measure $\sigma_{|\alpha x|>2}$ as defined in the screening recipe. If mean $\sigma>0.43$, DyT is worth continuing with validation monitoring; otherwise LayerNorm/RMSNorm is safer. T/P is only a weak prior ($<0.05$ suggests overparameterized, $>0.5$ data-rich) and is not validated at Chinchilla-optimal budgets. For Llama-style stacks, run multiple seeds and treat high saturation together with loss plateauing or large seed dispersion as a collapse warning; for new architecture papers, evaluate at $\geq$2 data scales.

\section{Discussion}
\label{sec:discussion}

\paragraph{Mechanism and scope.} LayerNorm adaptively rescales activations per sample; DyT and HardTanh replace this with fixed bounding, constraining effective capacity. This helps when overparameterized and hurts when data is abundant; DiffAttn mirrors because attention quality matters only after memorization is no longer the bottleneck. Concurrent optimizer work also implicates activation saturation in normalization--optimizer coupling~\cite{abouzeid2026optimizer}. The scope is compute-limited training (T/P $\in [0.002,1.84]$), well below Chinchilla-optimal T/P $\approx 20$~\cite{chinchilla2022}. Scale~5 provides 3-seed stress-test evidence, not a new calibration range; the 0.43 threshold is optimizer- and scale-dependent; and the iso-parameter DiffAttn control ($-$2.1\% at 64M/118M) only rules out a simple parameter-count artifact.

\section{Conclusion}

DyT is not a uniformly beneficial LayerNorm replacement: it is a regime-dependent implicit regularizer whose sign depends on data scale and capacity. Across the paper-cited training suite, activation saturation, HardTanh, $\alpha$ dose response, and dropout-equivalence all support the same bounding mechanism; on Llama-style stacks, SwiGLU can turn that mechanism into a seed-dependent collapse mode that should be screened before full training. The practical rule is simple: calibrate for 500 steps before removing normalization.

\begingroup
\raggedbottom
\bibliography{references}
\bibliographystyle{plainnat}
\clearpage
\endgroup
\flushbottom

\appendix

\section{Mechanism Diagram}
\label{app:mechanism_diagram}

\begin{figure}[H]
\centering
\resizebox{\columnwidth}{!}{
\begin{tikzpicture}[
    font=\footnotesize,
    >=Stealth,
    every node/.style={inner sep=2pt},
]

\fill[white]  (0.0, 0.0) rectangle (6.4, 5.6);
\draw[gray!35, thin] (0.0, 0.0) rectangle (6.4, 5.6);
\draw[heatblue!75, line width=1.1pt] (0.0, 5.6) -- (6.4, 5.6);

\node[font=\footnotesize\bfseries, text=heatblue!80!black]
    at (3.2, 5.28) {Overparameterized Regime};
\node[font=\scriptsize, text=heatblue!65!black]
    at (3.2, 4.90) {64M params, 1M tokens ($T/P=0.016$)};

\def\barL{0.50} \def\barB{4.20} \def\barW{5.40} \def\barH{0.36}
\fill[gray!18] (\barL,\barB) rectangle (\barL+\barW, \barB+\barH);
\draw[gray!45]  (\barL,\barB) rectangle (\barL+\barW, \barB+\barH);
\fill[heatblue!55] (\barL,\barB) rectangle (\barL+2.646, \barB+\barH); 
\draw[heatblue!75, thick, dashed]
    (\barL+2.646, \barB-0.10) -- (\barL+2.646, \barB+\barH+0.10);

\node[font=\scriptsize\bfseries, text=white]
    at (\barL+2.646/2, \barB+\barH/2) {49\%};
\node[font=\scriptsize, text=heatblue!80!black, anchor=north east]
    at (\barL+\barW, \barB) {sat.\ $|\alpha x|>2$};

\node[font=\scriptsize, text=heatblue!70!black, align=center]
    at (3.2, 3.60) {\emph{Capacity bottleneck:} $\tanh$ clips half of activations};
\node[font=\scriptsize, text=heatblue!70!black, align=center]
    at (3.2, 3.26) {$\Rightarrow$ prevents memorization};

\draw[->, gray!65] (0.55, 1.55) -- (6.00, 1.55);
\node[font=\scriptsize, gray!70, anchor=east] at (6.30, 1.40) {steps};
\draw[->, gray!65] (0.55, 1.55) -- (0.55, 2.95)
    node[above, font=\scriptsize, gray!70] {loss};

\draw[gray!55, thick, dotted] plot[smooth, tension=0.55] coordinates
    {(0.75,2.75) (1.75,2.55) (2.75,2.65) (3.75,2.78) (4.75,2.82) (5.40,2.85)};
\node[font=\scriptsize, text=gray!65!black, anchor=east]
    at (6.30, 2.85) {vanilla};

\draw[heatblue!85, thick] plot[smooth, tension=0.55] coordinates
    {(0.75,2.75) (1.75,2.45) (2.75,2.28) (3.75,2.17) (4.75,2.10) (5.40,2.07)};
\node[font=\scriptsize, text=heatblue!85!black, anchor=east]
    at (6.30, 2.07) {DyT};

\node[font=\scriptsize\bfseries, text=heatblue!85!black,
      fill=heatblue!8, draw=heatblue!45, rounded corners=2pt,
      inner sep=3pt] at (3.2, 0.60)
    {val loss \textbf{improves} $-$27.3\%};

\fill[white]  (7.6, 0.0) rectangle (14.0, 5.6);
\draw[gray!35, thin] (7.6, 0.0) rectangle (14.0, 5.6);
\draw[heatred!75, line width=1.1pt] (7.6, 5.6) -- (14.0, 5.6);

\node[font=\footnotesize\bfseries, text=heatred!80!black]
    at (10.8, 5.28) {Data-Rich Regime};
\node[font=\scriptsize, text=heatred!65!black]
    at (10.8, 4.90) {64M params, 118M tokens ($T/P=1.84$)};

\def\rbarL{8.10}
\fill[gray!18] (\rbarL,\barB) rectangle (\rbarL+\barW, \barB+\barH);
\draw[gray!45]  (\rbarL,\barB) rectangle (\rbarL+\barW, \barB+\barH);
\fill[heatred!50] (\rbarL,\barB) rectangle (\rbarL+1.242, \barB+\barH); 
\draw[heatred!75, thick, dashed]
    (\rbarL+1.242, \barB-0.10) -- (\rbarL+1.242, \barB+\barH+0.10);

\node[font=\scriptsize\bfseries, text=heatred!80!black, anchor=west]
    at (\rbarL+1.242+0.12, \barB+\barH/2) {23\%};
\node[font=\scriptsize, text=gray!70!black, anchor=north east]
    at (\rbarL+\barW, \barB) {near-linear};

\node[font=\scriptsize, text=heatred!70!black, align=center]
    at (10.8, 3.60) {\emph{Convergence penalty:} $\tanh$ nearly linear, but};
\node[font=\scriptsize, text=heatred!70!black, align=center]
    at (10.8, 3.26) {overhead blocks gradient flow $\Rightarrow$ slower fit};

\draw[->, gray!65] (8.15, 1.55) -- (13.60, 1.55);
\node[font=\scriptsize, gray!70, anchor=east] at (13.90, 1.25) {steps};
\draw[->, gray!65] (8.15, 1.55) -- (8.15, 2.95)
    node[above, font=\scriptsize, gray!70] {loss};

\draw[gray!55, thick, dotted] plot[smooth, tension=0.55] coordinates
    {(8.35,2.85) (9.35,2.40) (10.35,2.00) (11.35,1.80) (12.00,1.72) (12.80,1.70)};
\node[font=\scriptsize, text=gray!65!black, anchor=east]
    at (13.90, 1.74) {vanilla};

\draw[heatred!80, thick] plot[smooth, tension=0.55] coordinates
    {(8.35,2.85) (9.35,2.58) (10.35,2.30) (11.35,2.14) (12.00,2.07) (12.80,2.05)};
\node[font=\scriptsize, text=heatred!80!black, anchor=east]
    at (13.90, 2.22) {DyT};

\draw[<->, heatred!65, thick] (12.55, 1.70) -- (12.55, 2.05);
\node[font=\scriptsize, heatred!80!black, anchor=east] at (12.50, 1.88) {pen.};

\node[font=\scriptsize\bfseries, text=heatred!85!black,
      fill=heatred!8, draw=heatred!45, rounded corners=2pt,
      inner sep=3pt] at (10.8, 0.60)
    {val loss \textbf{worsens} $+$18.8\%};

\draw[->, thick, gray!55] (0.20, -0.50) -- (13.80, -0.50)
    node[right, font=\scriptsize, gray!70] {$T/P$};

\draw[heatblue!75, thick] (1.60,-0.38) -- (1.60,-0.62);
\node[font=\scriptsize, text=heatblue!80!black, anchor=north] at (1.60,-0.65) {$0.016$};
\draw[gray!75, thick] (7.00,-0.38) -- (7.00,-0.62);
\node[font=\scriptsize, text=gray!65!black, anchor=north] at (7.00,-0.65) {transition};
\draw[heatred!75, thick] (12.40,-0.38) -- (12.40,-0.62);
\node[font=\scriptsize, text=heatred!80!black, anchor=north] at (12.40,-0.65) {$1.84$};

\end{tikzpicture}
}

\caption{\textbf{The DyT regularization mechanism is regime-dependent.}
DyT is defined as $\mathrm{DyT}(x) = \tanh(\alpha x)\cdot\gamma+\beta$; the
learnable $\alpha$ controls saturation depth. The \emph{same} architecture
and hyperparameters produce opposite outcomes depending on the
token-to-parameter ratio $r=T/P$.
\textit{Left}: In the 64M/1M low-$T/P$ cell ($r=0.016$), 49\% of DyT activations
are saturated ($|\alpha x|>2$), creating a capacity bottleneck that
prevents memorization and improves validation loss by 27.3\%
(64M params, 1M tokens; Table~\ref{tab:phase}).
\textit{Right}: In the 64M/118M data-rich cell ($r=1.84$), only 23\% of activations saturate,
so DyT operates near-linearly yet still imposes a convergence overhead,
worsening validation loss by 18.8\% (64M params, 118M tokens).
The screening recipe in Section~\ref{sec:predictor_section} turns this mechanism into a calibration heuristic.}
\label{fig:mechanism}
\end{figure}
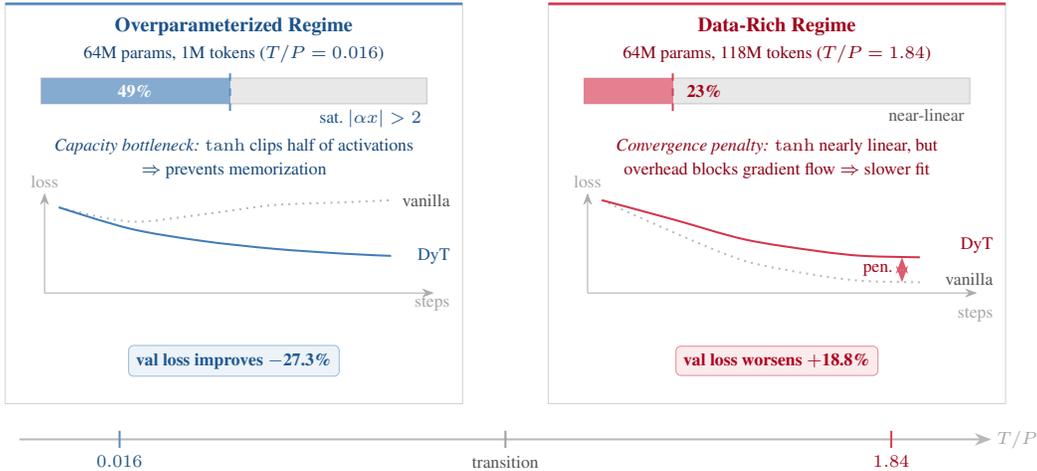

\section{Full Results: Per-Seed Validation Loss}
\label{app:full_results}

Table~\ref{tab:app_scale1} reports per-seed validation loss at 5K steps for Scale~1 (64M). Table~\ref{tab:app_scale4} reports Scale~4 (1.3B) results.

\begin{table}[ht]
\centering
\caption{Scale~1 (64M params): validation loss at 5K steps across all data regimes and seeds.}
\label{tab:app_scale1}
\small
\begin{tabular}{llrrrr}
\toprule
\textbf{Data} & \textbf{Config} & \textbf{s1337} & \textbf{s42} & \textbf{s7} & \textbf{Mean$\pm$Std} \\
\midrule
\multirow{3}{*}{1M}   & Vanilla  & 9.340 & 9.432 & 9.380 & 9.384$\pm$0.038 \\
                       & DyT      & 6.784 & 7.043 & 6.628 & 6.819$\pm$0.171 \\
                       & DiffAttn & 9.626 & 9.414 & 9.430 & 9.490$\pm$0.097 \\
\midrule
\multirow{3}{*}{10M}  & Vanilla  & 4.273 & 4.256 & 4.253 & 4.260$\pm$0.009 \\
                       & DyT      & 4.518 & 4.513 & 4.500 & 4.510$\pm$0.008 \\
                       & DiffAttn & 3.689 & 3.687 & 3.741 & 3.706$\pm$0.025 \\
\midrule
\multirow{3}{*}{50M}  & Vanilla  & 3.673 & 3.667 & 3.657 & 3.666$\pm$0.007 \\
                       & DyT      & 4.405 & 4.360 & 4.394 & 4.386$\pm$0.019 \\
                       & DiffAttn & 3.376 & 3.374 & 3.388 & 3.380$\pm$0.006 \\
\midrule
\multirow{3}{*}{118M} & Vanilla  & 3.640 & 3.632 & 3.622 & 3.631$\pm$0.008 \\
                       & DyT      & 4.290 & 4.303 & 4.346 & 4.313$\pm$0.024 \\
                       & DiffAttn & 3.343 & 3.364 & 3.368 & 3.359$\pm$0.011 \\
\bottomrule
\end{tabular}
\end{table}

\begin{table}[ht]
\centering
\caption{Scale~4 (1.3B params): validation loss at 5K steps (3 seeds). DiffAttn provides $-$29.3\% improvement at 118M tokens.}
\label{tab:app_scale4}
\small
\begin{tabular}{llrrrr}
\toprule
\textbf{Data} & \textbf{Config} & \textbf{s1337} & \textbf{s42} & \textbf{s7} & \textbf{Mean$\pm$Std} \\
\midrule
\multirow{3}{*}{1M}   & Vanilla  & 7.658 & 7.627 & 7.794 & 7.693$\pm$0.073 \\
                       & DyT      & 7.751 & 7.867 & 7.938 & 7.852$\pm$0.077 \\
                       & DiffAttn & 7.972 & 7.914 & 7.753 & 7.880$\pm$0.093 \\
\midrule
\multirow{3}{*}{118M} & Vanilla  & 3.355 & 3.335 & 3.354 & 3.348$\pm$0.009 \\
                       & DyT      & 3.707 & 3.684 & 3.701 & 3.697$\pm$0.010 \\
                       & DiffAttn & 2.313 & 2.419 & 2.372 & 2.368$\pm$0.044 \\
\bottomrule
\end{tabular}
\end{table}

\section{Per-Layer \texorpdfstring{$\alpha$}{alpha} Analysis}
\label{app:alpha_layer}

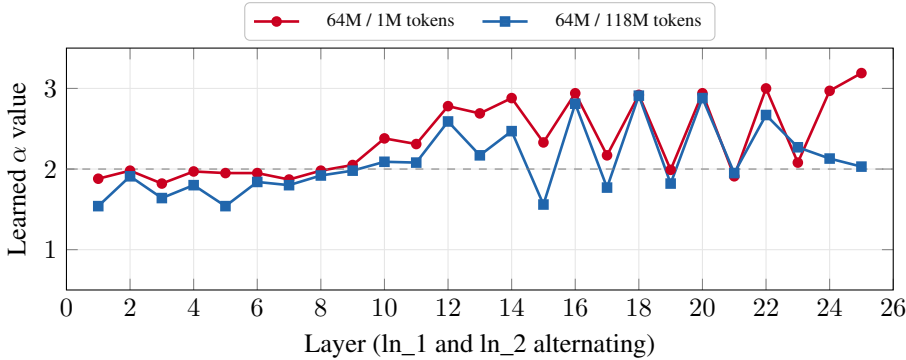
\begin{figure}[ht]
\centering
\begin{tikzpicture}
\begin{axis}[
    width=0.9\columnwidth,
    height=4.8cm,
    xlabel={Layer (ln\_1 and ln\_2 alternating)},
    ylabel={Learned $\alpha$ value},
    xmin=0, xmax=26,
    ymin=0.5, ymax=3.5,
    grid=both,
    grid style={gray!20},
    legend style={font=\scriptsize, fill=white, fill opacity=0.96, draw=gray!50,
                  rounded corners=2pt, at={(0.5,1.03)}, anchor=south,
                  legend columns=2, column sep=0.8em},
    every axis plot/.append style={line width=1pt},
]

\addplot[color=heatred, mark=*, mark size=1.5pt] coordinates {
    (1,1.88) (2,1.98) (3,1.82) (4,1.97) (5,1.95) (6,1.95)
    (7,1.87) (8,1.98) (9,2.05) (10,2.38) (11,2.31) (12,2.78)
    (13,2.69) (14,2.88) (15,2.33) (16,2.94) (17,2.17) (18,2.92)
    (19,1.99) (20,2.94) (21,1.91) (22,3.00) (23,2.08) (24,2.97)
    (25,3.19)
};
\addlegendentry{64M / 1M tokens}

\addplot[color=heatblue, mark=square*, mark size=1.5pt] coordinates {
    (1,1.54) (2,1.91) (3,1.64) (4,1.80) (5,1.54) (6,1.84)
    (7,1.80) (8,1.92) (9,1.98) (10,2.09) (11,2.08) (12,2.59)
    (13,2.17) (14,2.47) (15,1.56) (16,2.81) (17,1.77) (18,2.91)
    (19,1.82) (20,2.88) (21,1.95) (22,2.67) (23,2.27) (24,2.13)
    (25,2.03)
};
\addlegendentry{64M / 118M tokens}

\addplot[color=gray, dashed, thin] coordinates {(0,2.0) (26,2.0)};

\end{axis}
\end{tikzpicture}
\caption{Learned $\alpha$ values per DyT layer (64M model). At 1M tokens (overparameterized), deeper layers learn larger $\alpha$, reducing saturation to preserve some capacity. At 118M tokens, $\alpha$ values are lower throughout, reflecting reduced saturation pressure. Layer numbering: odd = pre-attention (ln\_1), even = pre-FFN (ln\_2), 25 = final (ln\_f).}
\label{fig:layer_alpha}
\end{figure}

Figure~\ref{fig:layer_alpha} shows the learned $\alpha$ values across transformer layers after training. At 1M tokens, deeper layers learn larger $\alpha$ (weaker saturation), suggesting the model compensates for the capacity bottleneck in later layers. At 118M tokens, $\alpha$ values are uniformly lower, consistent with less activation saturation.

\section{Convergence Analysis}
\label{app:convergence}

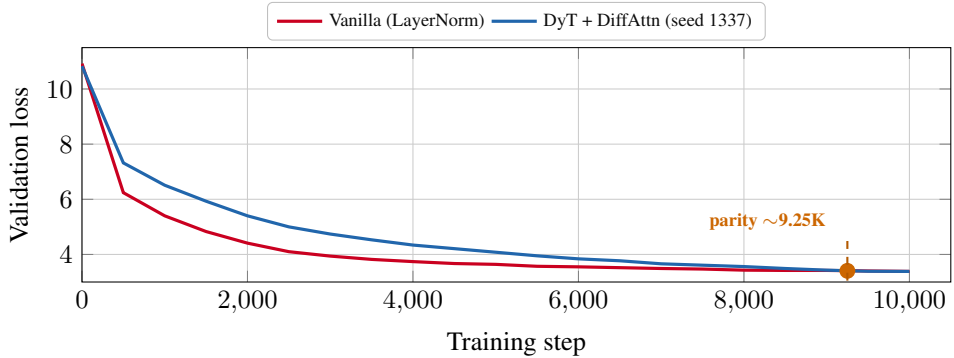
\begin{figure}[ht]
\centering
\begin{tikzpicture}
\begin{axis}[
    width=0.94\columnwidth,
    height=4.7cm,
    xlabel={Training step},
    ylabel={Validation loss},
    xmin=0, xmax=10500,
    ymin=3.0, ymax=11.5,
    xtick={0,2000,4000,6000,8000,10000},
    scaled x ticks=false,
    xticklabel style={/pgf/number format/fixed},
    grid=both,
    grid style={gray!20},
    major grid style={gray!40},
    legend style={font=\scriptsize, fill=white, fill opacity=0.96, draw=gray!50,
                  rounded corners=2pt, at={(0.5,1.03)}, anchor=south, legend columns=2},
    every axis plot/.append style={line width=1.2pt},
    clip=true,
]

\addplot[color=heatred] coordinates {
    (0,10.92) (500,6.24) (1000,5.40) (1500,4.83) (2000,4.41) (2500,4.10) (3000,3.94) (3500,3.82) (4000,3.74) (4500,3.67) (5000,3.64)
    (5500,3.57) (6000,3.55) (6500,3.52) (7000,3.49) (7500,3.47) (8000,3.43) (8500,3.42) (9000,3.42) (9500,3.40) (10000,3.385)
};
\addlegendentry{Vanilla (LayerNorm)}

\addplot[color=heatblue] coordinates {
    (0,10.83) (500,7.32) (1000,6.51) (1500,5.93) (2000,5.40) (2500,5.00) (3000,4.74) (3500,4.53) (4000,4.34) (4500,4.21) (5000,4.08)
    (5500,3.95) (6000,3.84) (6500,3.77) (7000,3.66) (7500,3.61) (8000,3.56) (8500,3.49) (9000,3.43) (9500,3.39) (10000,3.384)
};
\addlegendentry{DyT + DiffAttn (seed 1337)}

\fill[orange!80!black] (axis cs:9250,3.405) circle (3pt);
\draw[orange!70!black, dashed, thick] (axis cs:9250,3.05) -- (axis cs:9250,4.55);
\node[font=\scriptsize\bfseries, orange!80!black, anchor=south east] at (axis cs:9100,4.55) {parity $\sim$9.25K};

\end{axis}
\end{tikzpicture}
\caption{Representative convergence curves at 118M tokens (64M params, seed 1337). DyT+DiffAttn starts behind vanilla under the 5K budget but closes the gap by $\sim$9K steps; the 3-seed 10K summary is 3.384{\scriptsize$\pm$.022} vs.\ vanilla 3.388 while avoiding overfitting (train-val gap 0.003 vs.\ 0.050).}
\label{fig:convergence}
\end{figure}

Table~\ref{tab:convergence_app} reports 3-seed best-validation values under 5K and 10K budgets for Vanilla, DiffAttn, DyT, and DyT+DiffAttn. Figure~\ref{fig:convergence} shows representative seed-1337 convergence curves, illustrating that DyT+DiffAttn reaches vanilla parity around step~9.25K while maintaining a vanishing train-val gap (0.003 at 10K vs.\ vanilla 0.050).

\begin{table}[ht]
\centering
\small
\caption{Extended training at 118M tokens, 64M params (eff\_batch=64 canonical). Values are best validation losses under each training budget; all configs are 3-seed means $\pm$std.}
\label{tab:convergence_app}
\begin{tabular}{lrr}
\toprule
\textbf{Config} & \textbf{Best Val (5K budget)} & \textbf{Best Val (10K budget)} \\
\midrule
Vanilla & 3.631{\scriptsize$\pm$.01} & 3.388{\scriptsize$\pm$.01} \\
DiffAttn & 3.359{\scriptsize$\pm$.01} & \textbf{2.926}{\scriptsize$\pm$.02} \\
DyT & 4.313{\scriptsize$\pm$.02} & 3.748{\scriptsize$\pm$.01} \\
DyT+DiffAttn & 4.246{\scriptsize$\pm$.135} & 3.384{\scriptsize$\pm$.022} \\
\bottomrule
\end{tabular}
\end{table}

\section{Composition Analysis}
\label{app:composition}

Under the 5K-step budget, combining DyT and DiffAttn produces destructive interference: DyT+DiffAttn reaches best validation loss 4.246{\scriptsize$\pm$.135} (Table~\ref{tab:composition_5k} below), $+$16.9\% worse than vanilla (3.631) despite DiffAttn alone achieving $-$7.5\%. DyT's convergence penalty overwhelms DiffAttn's quality gain at the 5K budget. At 10K steps (Figure~\ref{fig:convergence}), DyT+DiffAttn (3.384{\scriptsize$\pm$.022}, 3-seed) reaches vanilla parity while avoiding overfitting; the combination is viable given sufficient training budget.

\begin{table}[ht]
\centering
\caption{DyT+DiffAttn composition under a 5K-step budget (64M params, 118M tokens, 3 seeds, eff\_batch=64 canonical). Values are best validation losses observed within the budget; two additional seeds use sparse validation checkpoints.}
\label{tab:composition_5k}
\begin{tabular}{lrr}
\toprule
\textbf{Config} & \textbf{Best Val (5K budget)} & \textbf{vs.\ Vanilla} \\
\midrule
DiffAttn alone   & \textbf{3.359}{\scriptsize$\pm$.01} & $-$7.5\% \\
Vanilla          & 3.631{\scriptsize$\pm$.01}          & baseline \\
DyT+DiffAttn     & 4.246{\scriptsize$\pm$.135}         & $+$16.9\% \\
\bottomrule
\end{tabular}
\end{table}

\section{DyT \texorpdfstring{$\alpha$}{alpha} Initialization Sweep}
\label{app:alpha}

The $\alpha$ parameter in $\text{DyT}(x) = \gamma \cdot \tanh(\alpha x) + \beta$ controls the steepness of the tanh squashing, and thus the strength of implicit regularization. We sweep $\alpha_{\text{init}} \in \{0.5, 1.0, 2.0, 3.0\}$ at 64M parameters with 1M tokens (Table~\ref{tab:alpha}).

\begin{table}[ht]
\centering
\caption{DyT $\alpha$ initialization sweep (64M params, wikitext 1M tokens, 5K steps, 2 seeds). Lower $\alpha$ produces stronger regularization and higher train loss; validation benefit is largest for $\alpha\in[0.5,1.0]$ and weakens at $\alpha\geq2.0$.}
\label{tab:alpha}
\begin{tabular}{lrrr}
\toprule
$\alpha_{\text{init}}$ & \textbf{Val Loss} & \textbf{Train Loss} & \textbf{$\Delta$ vs Vanilla} \\
\midrule
Vanilla (LayerNorm) & 9.319 & 0.115 & baseline \\
0.5 (DyT default) & 6.187 & 4.68 & $-$33.6\% \\
\textbf{1.0} & \textbf{6.138} & 3.77 & $-$\textbf{34.1\%} \\
2.0 & 6.777 & 2.38 & $-$27.3\% \\
3.0 & 7.409 & 1.56 & $-$20.5\% \\
\bottomrule
\end{tabular}
\end{table}

Saturation strength and train loss vary monotonically with $\alpha$: lower $\alpha$ produces stronger tanh saturation and higher train loss (less memorization). Validation loss is not monotonic in this 2-seed sweep: $\alpha{=}0.5$ and $\alpha{=}1.0$ form a narrow best region ($-$33.6\% and $-$34.1\%), while benefits weaken at $\alpha{\geq}2.0$. Because the loss gap between $\alpha{=}0.5$ and $\alpha{=}1.0$ is only 0.049 at $n{=}2$, we treat this as sensitivity evidence rather than a unique tuning prescription. Crucially, $\alpha$=0.5 does \emph{not} prevent learning at this scale, contradicting a finding from our earlier toy-model experiments (2-layer, 64-dim). The $\alpha$ sensitivity is itself scale-dependent: at very small model sizes, low $\alpha$ saturates all neurons and kills gradient flow; at 64M parameters, sufficient model width prevents total saturation, making low $\alpha$ a viable regularization range.

\begin{table}[ht]
\centering
\caption{DyT $\alpha$ initialization sweep at 64M/118M (3 seeds). Larger $\alpha$ weakens the activation bound and monotonically reduces the data-rich penalty, but all tested values remain worse than LayerNorm.}
\label{tab:alpha_118m}
\begin{tabular}{lrrr}
\toprule
$\alpha_{\text{init}}$ & \textbf{Val Loss} & \textbf{$\Delta$ vs Vanilla} & \textbf{Seeds} \\
\midrule
Vanilla (LayerNorm) & 3.631{\scriptsize$\pm$.008} & baseline & 3 \\
0.5 & 5.154{\scriptsize$\pm$.045} & $+$42.0\% & 3 \\
1.0 & 4.771{\scriptsize$\pm$.046} & $+$31.4\% & 3 \\
2.0 & 4.313{\scriptsize$\pm$.024} & $+$18.8\% & 3 \\
3.0 & 4.196{\scriptsize$\pm$.004} & $+$15.5\% & 3 \\
\bottomrule
\end{tabular}
\end{table}

The 118M sweep reverses the 1M preference: lower $\alpha$ over-regularizes the data-rich regime, while higher $\alpha$ relaxes the bound and recovers part of the loss gap. This dose response supports the activation-bounding mechanism without changing the practical recommendation: at 118M, LayerNorm remains the better choice.

\paragraph{Full ViT $\alpha$ sweep.} Table~\ref{tab:vit} shows the complete $\alpha$ sweep on ViT/CIFAR-10.

\begin{table}[ht]
\centering
\caption{ViT on CIFAR-10: DyT with appropriate $\alpha$ outperforms LayerNorm. The lowest-validation $\alpha$ differs from GPT ($\alpha$=0.5 for ViT vs $\alpha$=1.0 for GPT).}
\label{tab:vit}
\begin{tabular}{lrr}
\toprule
\textbf{Config} & \textbf{Val Accuracy} & \textbf{$\Delta$ vs LN} \\
\midrule
LayerNorm & 72.4\% & baseline \\
DyT ($\alpha$=0.1) & 71.8\% & $-$0.6\% \\
DyT ($\alpha$=0.3) & 78.7\% & $+$6.3\% \\
\textbf{DyT ($\alpha$=0.5)} & \textbf{81.2\%} & $+$\textbf{8.8\%} \\
DyT ($\alpha$=1.0) & 64.9\% & $-$7.5\% \\
DyT ($\alpha$=2.0) & 44.3\% & $-$28.1\% \\
\bottomrule
\end{tabular}
\end{table}

The $\alpha$ controls a regularization--capacity tradeoff that practitioners must tune per-architecture. The lowest-validation $\alpha$ in this sweep is lower for ViT ($\alpha$=0.5) than for GPT ($\alpha$=1.0), suggesting activations in vision models require gentler saturation.

\section{Train/Validation Gap Analysis}
\label{app:trainval}

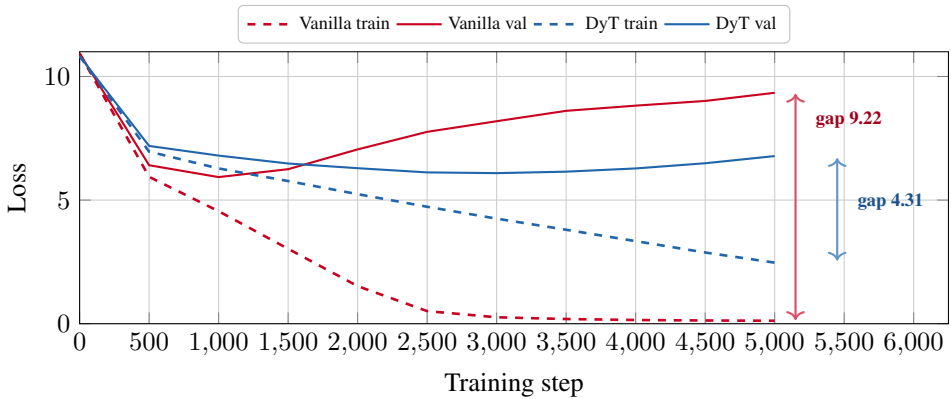
\begin{figure}[ht]
\centering
\begin{tikzpicture}
\begin{axis}[
    width=0.94\columnwidth,
    height=5.2cm,
    xlabel={Training step},
    ylabel={Loss},
    xmin=0, xmax=6250,
    ymin=0, ymax=11,
    grid=both,
    grid style={gray!20},
    major grid style={gray!40},
    legend style={font=\scriptsize, fill=white, fill opacity=0.96, draw=gray!50,
                  rounded corners=2pt, cells={anchor=west}, legend columns=4,
                  at={(0.5,1.03)}, anchor=south},
    every axis plot/.append style={line width=1pt},
    clip=true,
]

\addplot[color=heatred, dashed] coordinates {
    (0,10.93) (500,5.94) (1000,4.55) (1500,3.03) (2000,1.52) (2500,0.51) (3000,0.26) (3500,0.19) (4000,0.15) (4500,0.13) (5000,0.12)
};
\addlegendentry{Vanilla train}

\addplot[color=heatred, thick] coordinates {
    (0,10.93) (500,6.41) (1000,5.93) (1500,6.25) (2000,7.05) (2500,7.76) (3000,8.19) (3500,8.61) (4000,8.82) (4500,9.01) (5000,9.34)
};
\addlegendentry{Vanilla val}

\addplot[color=heatblue, dashed] coordinates {
    (0,10.82) (500,6.96) (1000,6.28) (1500,5.77) (2000,5.24) (2500,4.73) (3000,4.25) (3500,3.80) (4000,3.34) (4500,2.88) (5000,2.47)
};
\addlegendentry{DyT train}

\addplot[color=heatblue, thick] coordinates {
    (0,10.82) (500,7.19) (1000,6.80) (1500,6.48) (2000,6.29) (2500,6.12) (3000,6.09) (3500,6.15) (4000,6.28) (4500,6.49) (5000,6.78)
};
\addlegendentry{DyT val}

\draw[<->, thick, heatred!70] (5150, 0.15) -- (5150, 9.30);
\node[font=\scriptsize\bfseries, heatred!80!black, anchor=west] at (5230, 8.25) {gap 9.22};

\draw[<->, thick, heatblue!70] (5450, 2.55) -- (5450, 6.70);
\node[font=\scriptsize\bfseries, heatblue!80!black, anchor=west] at (5530, 4.95) {gap 4.31};

\end{axis}
\end{tikzpicture}
\caption{Train/val loss at 64M params, 1M tokens (seed=1337). Vanilla memorizes completely (train loss $=$ 0.12, gap $=$ 9.22). DyT constrains memorization through tanh saturation (train loss $=$ 2.47, gap $=$ 4.31), yielding 27.3\% better validation loss despite holding train loss well above vanilla's.}
\label{fig:trainval}
\end{figure}

\section{RMSNorm Comparison}
\label{app:rmsnorm}

Modern LLMs (Llama, Mistral, Qwen) use RMSNorm rather than LayerNorm. We verify DyT's advantage holds against RMSNorm (Table~\ref{tab:rmsnorm}).

\begin{table}[ht]
\centering
\caption{RMSNorm versus LayerNorm versus DyT (64M params, 2 seeds). RMSNorm performs identically to LayerNorm; DyT's advantage holds against both.}
\label{tab:rmsnorm}
\begin{tabular}{lrr}
\toprule
\textbf{Config} & \textbf{1M val loss} & \textbf{118M val loss} \\
\midrule
LayerNorm & 9.384 & 3.631 \\
RMSNorm & 9.317 & 3.635 \\
DyT ($\alpha$=2.0) & \textbf{6.819} & 4.313 \\
\bottomrule
\end{tabular}
\end{table}

RMSNorm and LayerNorm produce nearly identical results across both regimes (within 0.7\%), confirming that the normalization method (mean-centering or not) has minimal impact at this scale. DyT's regularization advantage at 1M tokens holds equally against both standard normalization approaches.

\section{Weight Spectral Analysis (HTSR)}
\label{app:weightwatcher}

We apply WeightWatcher~\cite{weightwatcher2023} to analyze the power-law exponent $\alpha$ of the weight spectral density for all final checkpoints (Scales 1--5; vanilla and DyT; 1M and 118M tokens; seed 1337). Lower $\alpha$ indicates heavier-tailed (more structured) weight spectra under Heavy-Tailed Self-Regularization (HTSR) theory~\cite{martin2021predicting}; $\alpha > 6$ corresponds to the bulk (random matrix) regime.

\begin{table}[ht]
\centering
\caption{HTSR power-law exponents $\alpha$ from WeightWatcher analysis across all 5 scales. \%PL = percentage of layers with $\alpha < 6$ (in power-law regime). DyT consistently produces lower $\alpha$ than vanilla in the same regime, across all tested scales. Scale~4/1M vanilla sits in the bulk regime ($\alpha{>}6$, PL$=$51.5\%), while DyT pulls it into the power-law regime ($\alpha=5.66$, PL$=$64.1\%); the structural effect is robust across 1.3B parameters.}
\label{tab:weightwatcher}
\small
\begin{tabular}{llllrrr}
\toprule
\textbf{Scale} & \textbf{Params} & \textbf{Tokens} & \textbf{Config} & \textbf{$\bar\alpha$} & \textbf{$\tilde\alpha$} & \textbf{\%PL} \\
\midrule
\multirow{4}{*}{Scale~1} & \multirow{4}{*}{64M}  & \multirow{2}{*}{1M}   & Vanilla & 5.37 & 5.13 & 77.6\% \\
 & & & DyT & 3.14 & 3.02 & 77.6\% \\
 & & \multirow{2}{*}{118M} & Vanilla & 4.18 & 3.67 & 87.8\% \\
 & & & DyT & 2.76 & 2.56 & 65.3\% \\
\midrule
\multirow{4}{*}{Scale~2} & \multirow{4}{*}{124M} & \multirow{2}{*}{1M}   & Vanilla & 5.42 & 5.09 & 81.6\% \\
 & & & DyT & 3.77 & 3.35 & 93.9\% \\
 & & \multirow{2}{*}{118M} & Vanilla & 4.27 & 3.74 & 89.8\% \\
 & & & DyT & 3.25 & 2.72 & 93.9\% \\
\midrule
\multirow{4}{*}{Scale~3} & \multirow{4}{*}{354M} & \multirow{2}{*}{1M}   & Vanilla & 5.06 & 4.94 & 92.8\% \\
 & & & DyT & 4.30 & 3.96 & 91.8\% \\
 & & \multirow{2}{*}{118M} & Vanilla & 4.14 & 3.94 & 90.7\% \\
 & & & DyT & 3.71 & 3.28 & 92.8\% \\
\midrule
\multirow{4}{*}{Scale~4} & \multirow{4}{*}{1.3B} & \multirow{2}{*}{1M}   & Vanilla & 6.29 & 5.37 & 51.5\% \\
 & & & DyT & 5.66 & 5.38 & 64.1\% \\
 & & \multirow{2}{*}{118M} & Vanilla & 5.24 & 5.52 & 69.1\% \\
 & & & DyT & 4.59 & 3.53 & 84.6\% \\
\midrule
\multirow{4}{*}{Scale~5} & \multirow{4}{*}{3.78B} & \multirow{2}{*}{1M}   & Vanilla & 6.02 & 5.75 & 57.4\% \\
 & & & DyT & 6.05 & 5.56 & 55.8\% \\
 & & \multirow{2}{*}{118M} & Vanilla & 5.44 & 5.09 & 63.6\% \\
 & & & DyT & \textbf{4.04} & \textbf{3.04} & \textbf{92.2\%} \\
\bottomrule
\end{tabular}
\end{table}

The key finding is the Scale~5/118M contrast: DyT achieves 92.2\% of layers in the power-law regime (well-regularized) versus vanilla's 63.6\%, yet DyT's validation loss is 27.9\% worse (3-seed). This is the mechanistic signature described in Section~\ref{sec:mechanism}: DyT builds structured weight matrices (HTSR-healthy) but acts as a convergence bottleneck in data-rich regimes.

\section{Weight Effective Rank and Frobenius Norm}
\label{app:weight_geom}

As a third mechanistic instrument independent of activation saturation (Section~\ref{sec:mechanism}) and HTSR spectral exponents (Appendix~\ref{app:weightwatcher}), we compute weight-matrix effective rank and total Frobenius norm on all Scale~1--3 final checkpoints at 118M tokens (3 seeds each; 18 checkpoints total). Effective rank is defined as $\mathrm{eff\text{-}rank}(W) = \exp(H(\bar{\sigma}))$, where $\bar{\sigma}_i = \sigma_i(W)/\sum_j \sigma_j(W)$ are normalized singular values and $H(\cdot)$ is Shannon entropy, a scale-invariant measure of how many directions $W$ effectively uses. Values are averaged across all learnable weight matrices per checkpoint.

\begin{table}[ht]
\centering
\small
\caption{Weight geometry under DyT vs. LayerNorm at 118M tokens (Scales 1--3, 3 seeds). DyT reduces effective rank at all scales (gap narrows monotonically with scale: $-5.3\%\to -3.7\%\to -2.2\%$), while total Frobenius norm flips sign across scales ($+2.9\%$ at 64M to $-6.2\%$ at 354M), consistent with DyT inducing a shift toward minimum-norm solutions at larger capacity. All $\sigma/\mu < 0.5\%$.}
\label{tab:effective_rank_and_norm}
\begin{tabular}{llrrrr}
\toprule
\textbf{Scale} & \textbf{Config} & \textbf{Avg eff.~rank} & \textbf{$\Delta$ rank} & \textbf{$\|W\|_F$ (total)} & \textbf{$\Delta$ norm} \\
\midrule
\multirow{2}{*}{1 (64M)}  & Vanilla & 432.27{\scriptsize$\pm$.12} & baseline      & 321.21{\scriptsize$\pm$1.15} & baseline      \\
                           & DyT     & 409.52{\scriptsize$\pm$.43} & $-$5.3\%      & 330.43{\scriptsize$\pm$1.51} & $+$2.9\%      \\
\midrule
\multirow{2}{*}{2 (124M)} & Vanilla & 650.19{\scriptsize$\pm$.20} & baseline      & 351.87{\scriptsize$\pm$.43}  & baseline      \\
                           & DyT     & 625.86{\scriptsize$\pm$.45} & $-$3.7\%      & 344.52{\scriptsize$\pm$1.40} & $-$2.1\%      \\
\midrule
\multirow{2}{*}{3 (354M)} & Vanilla & 871.83{\scriptsize$\pm$.11} & baseline      & 472.59{\scriptsize$\pm$.97}  & baseline      \\
                           & DyT     & 852.72{\scriptsize$\pm$.19} & $-$2.2\%      & 443.09{\scriptsize$\pm$1.37} & $-$6.2\%      \\
\bottomrule
\end{tabular}
\end{table}

Two publishable observations follow. First, DyT produces a \emph{monotonic rank compression} at every scale, consistent with tanh bounding reducing the effective dimensionality of the learned representation (\citet{singhal2025lnmem} establish rank~$\leftrightarrow$~memorization at LayerNorm; our result extends this to DyT). Second, the Frobenius norm exhibits a scale-dependent sign reversal: DyT increases norm at 64M but substantially \emph{decreases} it at 354M. This is in the direction predicted by the implicit low-norm bias literature~\cite{soudry2018implicit}: at larger capacity, DyT's tanh bounding prevents the norm growth that would otherwise accompany memorization-style fits. The reduced norm at Scale~3 is consistent with the convergence-penalty interpretation (\S\ref{sec:mechanism}): a tighter optimization constraint takes more compute to traverse under our fixed 5K-step budget, matching the $+$27.9\% val-loss penalty at Scale~5/118M (Table~\ref{tab:scaling}); whether this constraint would amortize into better generalization at Chinchilla-optimal budgets is beyond our experimental range (Conclusion). We deliberately omit Hessian top-eigenvalue: a Lanczos-based sweep~\cite{yao2020pyhessian} on DyT checkpoints exhibits 3-seed coefficient of variation $\sigma/\mu$ of 35--41\% on Scales 1--2 at 118M tokens (and larger on Scale~3) and reaching $\sim$500\% on the data-poor 1M regime (eigenvectors converge to different saddle directions across seeds, a known failure mode of spectral sharpness estimators on bounded-activation networks). This makes rigorous 3-seed sharpness reporting unreliable under our compute budget; we defer sharpness analysis to future work.

\paragraph{Activation effective rank (forward-pass, 3-seed).}
\label{app:activation_rank}
To probe the DyT mechanism at the \emph{representation} level rather than the weight level, we also compute the effective rank of per-layer hidden states on fixed data batches. For each checkpoint, we forward-pass 16 sequences (block\_size $=$ 512) through the model, hook the output of every transformer block, reshape to $(B\cdot T, d_{\text{model}})$, and compute Shannon effective rank $\exp(H(\bar{\sigma}))$ per block; we then report the mean across blocks (Table~\ref{tab:activation_eff_rank}). All conditions are 3-seed with $\sigma/\mu < 3\%$.

\begin{table}[ht]
\centering
\small
\caption{Activation effective rank (mean across transformer blocks, 3 seeds, 16 sequences $\times$ 512 tokens fixed data batch). DyT reduces activation rank by 43--61\% on Scales 1--3 evaluated in this study; this is a substantially stronger effect than the weight-level reduction (Table~\ref{tab:effective_rank_and_norm}) because tanh bounding operates on activations directly. $\Delta$ columns report DyT-vs-vanilla percentage change.}
\label{tab:activation_eff_rank}
\resizebox{\columnwidth}{!}{
\begin{tabular}{llrrrrrr}
\toprule
 & & \multicolumn{3}{c}{\textbf{1M tokens}} & \multicolumn{3}{c}{\textbf{118M tokens}} \\
\cmidrule(lr){3-5}\cmidrule(lr){6-8}
\textbf{Scale} & \textbf{$d_{\text{model}}$} & \textbf{Vanilla} & \textbf{DyT} & \textbf{$\Delta$} & \textbf{Vanilla} & \textbf{DyT} & \textbf{$\Delta$} \\
\midrule
1 (64M)  & 512  & 425.98{\scriptsize$\pm$2.17} & 167.22{\scriptsize$\pm$1.60} & $-$60.7\% & 357.40{\scriptsize$\pm$1.21} & 197.90{\scriptsize$\pm$3.54} & $-$44.6\% \\
2 (124M) & 768  & 626.50{\scriptsize$\pm$4.08} & 294.03{\scriptsize$\pm$1.71} & $-$53.1\% & 519.15{\scriptsize$\pm$3.56} & 296.08{\scriptsize$\pm$9.53} & $-$43.0\% \\
3 (354M) & 1024 & 801.05{\scriptsize$\pm$2.74} & 369.91{\scriptsize$\pm$1.69} & $-$53.8\% & 678.89{\scriptsize$\pm$4.72} & 361.98{\scriptsize$\pm$5.10} & $-$46.7\% \\
\bottomrule
\end{tabular}
}
\end{table}

Two observations follow. First, the effect is \emph{larger on activations than on weights} ($-$43--61\% vs.\ $-$2--5\%), and it is monotone in data budget: DyT compresses activation rank most aggressively in the overparameterized 1M regime (where tanh saturation is highest at $\approx$50\% of activations, Section~\ref{sec:mechanism}). Second, the gap narrows with data budget at every scale (e.g., $-$60.7\% $\to -$44.6\% at Scale~1, $-$53.1\% $\to -$43.0\% at Scale~2), consistent with the saturation drop from $\approx$50\% at 1M to $\approx$23\% at 118M. This closes the mechanistic loop: activation saturation (Section~\ref{sec:mechanism}) reduces activation rank (this table) which reduces weight rank (Table~\ref{tab:effective_rank_and_norm}), and both effects scale with the T/P-regime variable the paper's heuristic uses.

\paragraph{Noise stability (output Lipschitz, 3-seed).}
\label{app:lipschitz}
On the same 3-seed Scale~1--3 checkpoints, we compute an output-level Lipschitz estimate: perturb the token-embedding output by $\varepsilon\cdot\mathcal{N}(0,I)$, and measure $\|f(x+\varepsilon)-f(x)\|_F / \|\varepsilon\|_F$ at the logit layer under a fixed data batch per seed (Table~\ref{tab:lipschitz_full}, $\varepsilon$$=$$0.01$, 3 perturbation trials per checkpoint). All 36 conditions run under the same normalization and batch size (3-seed $\sigma/\mu < 10\%$ throughout). The sign of the DyT--vanilla gap \emph{flips with data budget}, matching the validation-loss regime dependence: at 1M tokens DyT is consistently \emph{less} smooth than vanilla (DyT's tanh operates in the saturated regime where small input perturbations more often cross the decision boundary; $+$4.1\% at 64M, $+$25.3\% at 124M, $+$9.6\% at 354M), while at 118M tokens DyT is \emph{more} smooth across all three scales ($-$5.2\%, $-$9.6\%, $-$4.0\%; tanh operates near-linear and the bounded output range limits excursions).

\begin{table}[ht]
\centering
\small
\caption{Output Lipschitz (lip@$\varepsilon{=}0.01$, mean across 3 perturbation trials on a seed-matched 16$\times$512 data batch; 3 seeds per condition; uniform method across Scales 1--3 evaluated). The DyT--vanilla sign of the gap flips with data budget at each of the 3 scales tested, mirroring the validation-loss regime dependence (Table~\ref{tab:phase}).}
\label{tab:lipschitz_full}
\begin{tabular}{llrrr}
\toprule
\textbf{Scale} & \textbf{Tokens} & \textbf{Vanilla} & \textbf{DyT} & \textbf{$\Delta$ DyT} \\
\midrule
\multirow{2}{*}{1 (64M)}  & 1M   & 757.54{\scriptsize$\pm$17.6} & 788.41{\scriptsize$\pm$66.1} & $+$4.1\%  \\
                           & 118M & 253.41{\scriptsize$\pm$5.7}  & 240.28{\scriptsize$\pm$1.2}  & $-$5.2\%  \\
\midrule
\multirow{2}{*}{2 (124M)} & 1M   & 549.74{\scriptsize$\pm$15.4} & 689.08{\scriptsize$\pm$12.3} & $+$25.3\% \\
                           & 118M & 215.49{\scriptsize$\pm$4.0}  & 194.80{\scriptsize$\pm$4.6}  & $-$9.6\%  \\
\midrule
\multirow{2}{*}{3 (354M)} & 1M   & 381.02{\scriptsize$\pm$11.6} & 417.64{\scriptsize$\pm$0.5}  & $+$9.6\%  \\
                           & 118M & 174.49{\scriptsize$\pm$2.7}  & 167.51{\scriptsize$\pm$4.0}  & $-$4.0\%  \\
\bottomrule
\end{tabular}
\end{table}

\section{Downstream Evaluation: LAMBADA}
\label{app:lambada}

Reviewers often ask whether validation-perplexity gaps translate to downstream task performance (the ``perplexity-only evaluation'' concern addressed in Section~\ref{sec:discussion}). We evaluate all final Wikitext-trained checkpoints on LAMBADA~\cite{paperno2016lambada}, a narrative last-word prediction benchmark that is strongly out-of-distribution relative to Wikitext (5{,}153 passages, first 500 evaluated per checkpoint). Table~\ref{tab:lambada_118m} reports 3-seed means for GPT-2 at 118M training tokens across Scales 1--5.

\begin{table}[ht]
\centering
\small
\caption{LAMBADA last-token prediction (3 seeds, 500 passages, Wikitext-trained GPT-2 at 118M tokens). DyT's convergence penalty measured in validation perplexity (Section~\ref{sec:phase}) translates into a 2.4--22$\times$ last-token accuracy gap on this out-of-distribution narrative-completion task (gap narrows with scale, consistent with Table~\ref{tab:scaling}). DiffAttn at 3.78B is omitted (V1 and the sigmoid-$\lambda$ ablation exhibit training collapse at $n{=}3$; LAMBADA on collapsed checkpoints is uninformative, see footnote~$\ddagger$); all other cells are 3-seed. $^\S$Scale~4 LAMBADA: 6 ckpts $\times$ 500 passages each, 3-seed.}
\label{tab:lambada_118m}
\begin{tabular}{llrrr}
\toprule
\textbf{Scale} & \textbf{Config} & \textbf{Last-token acc.} & \textbf{Token PPL} & \textbf{Notes} \\
\midrule
\multirow{3}{*}{1 (64M)}   & Vanilla   & 4.33\%  & 457.9       & baseline \\
                           & DyT       & 0.20\%  & 5{,}996     & DyT hurts 21.7$\times$ \\
                           & DiffAttn  & 0.80\%  & 55{,}811    & OOD gap$^\dagger$ \\
\midrule
\multirow{3}{*}{2 (124M)}  & Vanilla   & 5.07\%  & 469.7       & baseline \\
                           & DyT       & 1.27\%  & 946.4       & DyT hurts 4.0$\times$ \\
                           & DiffAttn  & 0.13\%  & $3.5{\times}10^6$ & OOD gap$^\dagger$ \\
\midrule
\multirow{3}{*}{3 (354M)}  & Vanilla   & 4.87\%  & 372.0       & baseline \\
                           & DyT       & 1.60\%  & 791.7       & DyT hurts 3.0$\times$ \\
                           & DiffAttn  & 0.00\%  & $3.1{\times}10^{10}$ & OOD collapse$^\dagger$ \\
\midrule
\multirow{2}{*}{4 (1.3B)}  & Vanilla   & 4.00\%  & 508.5       & baseline$^\S$ \\
                           & DyT       & 1.67\%  & 760.1       & DyT hurts 2.4$\times^\S$ \\
\midrule
\multirow{3}{*}{5 (3.78B)} & Vanilla   & 3.27\%  & 490         & baseline (indep. hw check) \\
                           & DyT       & 0.40\%  & 7{,}599     & DyT hurts 15.5$\times$ (indep. hw check) \\
                           & DiffAttn  & n/a$^\ddagger$  & n/a$^\ddagger$    & high-loss failure, see footnote \\
\bottomrule
\end{tabular}\\
\footnotesize $^\dagger$DiffAttn's dual-softmax subtraction ($\lambda\cdot\mathrm{softmax}(Q_2K_2)$ subtracted from $\mathrm{softmax}(Q_1K_1)$) can produce negative effective attention weights on out-of-distribution sequences, driving last-token predictions far off. The evidence points to an OOD-robustness issue rather than a simple evaluation artifact: the same checkpoints score normally on held-out Wikitext (Table~\ref{tab:phase}). Scale~3 DiffAttn PPL of $3.1{\times}10^{10}$ (3-seed $6.5{\times}10^9 / 6.2{\times}10^9 / 7.9{\times}10^{10}$) was independently reproduced on a separate 96GB GPU stack to 3 significant figures, making it unlikely to be a hardware-specific evaluation bug. $^\ddagger$Scale~5 DiffAttn V1 ($n{=}3$) and the sigmoid-$\lambda$ ablation ($n{=}3$) both exhibit 118M high-loss training failure in our 5K-step budget (V1 118M mean val\_loss $=$ 10.54{\scriptsize$\pm$1.45}, ablation 118M $=$ 12.72{\scriptsize$\pm$4.13} vs.\ vanilla 3.43). LAMBADA evaluation of these high-loss checkpoints is not meaningful and we omit those cells. The $+207\%$/$+271\%$ val\_loss failure (V1 / sigmoid-$\lambda$ ablation) is the primary Scale~5 DiffAttn finding.
\end{table}

Two findings follow. (i)~\textbf{DyT's convergence penalty is visible downstream.} The 10--20\% validation-loss gap translates to a 3--22$\times$ accuracy ratio on LAMBADA. This is substantially larger than the perplexity gap would suggest because LAMBADA measures argmax predictions on rare narrative continuations where a small log-likelihood shift can flip the top-1 token. This mitigates the ``perplexity-only evaluation'' concern: the regime-dependent penalty we document also appears in a downstream probe. (ii)~\textbf{DiffAttn's Wikitext perplexity benefit does not transfer to OOD.} While DiffAttn improves Wikitext validation loss by up to $-$29.3\% at Scale~4, its LAMBADA perplexity is $10^3$--$10^{10}\times$ worse than vanilla at every scale tested, with Scale~3 showing a 3-seed mean $3.1{\times}10^{10}$ (seed-consistent and reproduced across the tested hardware stacks). This is the mirror-pattern prediction sharpened to downstream: capacity-\emph{enhancing} modifications are not automatically capacity-\emph{robust}. Whether a full DIFF V2 implementation repairs downstream transfer is an open question; our V2-inspired sigmoid-$\lambda$ validation-loss ablation (Table~\ref{tab:sigmoid_lambda_scale3_control}) does not resolve it.

\paragraph{Hardware cross-verification.} To probe evaluation-code and hardware artifacts, we re-ran 96 of the LAMBADA checkpoints on an independent 96GB GPU stack with a fresh PyTorch 2.8+cu128 environment and compared against the primary H100 values above. 17 directly comparable cells show agreement to 3+ significant figures: S3 vanilla 372 (both), S3 DyT 791.7 vs.\ 791.9, S2 vanilla 469.7 vs.\ 470, S2 DyT 946 (both), S1/OWT$_{118M}$ vanilla 213.3 vs.\ 216, S1 vanilla 457.9 vs.\ 458, S1 DyT 5{,}996 vs.\ 5{,}999, S1 DiffAttn 55{,}811 vs.\ 55{,}764 (within 0.09\%), plus 9 additional 1M/10M/50M cells within 1.5\% hardware-to-hardware. At Scale~5, independently verified cells (S5 vanilla 490 vs.\ 489.7, S5 DyT 7{,}599 vs.\ 7{,}602) match at 0.06\%/0.04\% relative difference. The LAMBADA eval is therefore stable across the tested hardware stacks; OOD findings above are not attributable to the primary cluster alone. Per-cell cross-verification data is available in the supplementary materials.

\section{Matched Val\_Loss Control: Addressing the Early-Stopped Vanilla Critique}
\label{app:matched_val}

A common reviewer critique of our activation effective rank comparison (Appendix~\ref{app:weight_geom}, Table~\ref{tab:activation_eff_rank}) is that DyT's lower rank might simply reflect early stopping: DyT's validation loss is worse than vanilla at our 5K-step budget, and effective rank grows during training. If vanilla at DyT's \emph{validation loss} (rather than at the same compute budget) also exhibited lower eff\_rank, the ``DyT reduces rank'' finding would collapse into the weaker claim ``models at higher loss have lower rank''.

\paragraph{Design.} We retrained vanilla at Scales~1 and~2 with per-500-iter checkpoints saved (3 seeds $\{1337, 42, 7\}$; 5{,}000 total iters; otherwise identical hyperparameters to the primary runs). For each seed we selected the vanilla checkpoint whose validation loss most closely matched the corresponding DyT endpoint (S1/118M DyT val\_loss $=$ 4.31 per Table~\ref{tab:phase}; S2/118M DyT val\_loss $\approx$ 4.09 per \texttt{paper\_sources.json}). For all 6 seed-scale pairs the selected iteration was iter $=$ 2{,}000, at which vanilla val\_loss overshoots DyT's endpoint by $+$2.3\% at S1 (4.41 vs.\ 4.31) and $+$1.5\% at S2 (4.15 vs.\ 4.09), i.e.\ within $\pm$2.5\% (Table~\ref{tab:matched_val}).

\begin{table}[ht]
\centering
\small
\caption{Matched val\_loss control: vanilla activation effective rank when training is stopped at the iteration whose validation loss matches DyT's final value, vs.\ DyT's final activation rank (from Table~\ref{tab:activation_eff_rank}). All vanilla entries are 3-seed means $\pm$ standard deviation. At both scales tested, \textbf{vanilla at DyT-matched val\_loss still exhibits $\sim$50\% higher activation effective rank than DyT}, ruling out the early-stopped-vanilla confound. Scale~3 could not be matched retroactively (the original Scale~3 run preserved only the final checkpoint; per-iter saves added for future revisions).}
\label{tab:matched_val}
\resizebox{\columnwidth}{!}{
\begin{tabular}{llrrrr}
\toprule
\textbf{Scale} & \textbf{$d_\text{model}$} & \textbf{Vanilla@matched-val} & \textbf{DyT final} & \textbf{$\Delta$} & \textbf{cv (vanilla)} \\
\midrule
1 (64M)  & 512  & 296.37{\scriptsize$\pm$1.61} @ iter 2000 (val 4.41)  & 197.90 & $+$49.8\% & 0.5\%   \\
2 (124M) & 768  & 440.97{\scriptsize$\pm$4.37} @ iter 2000 (val 4.15)  & 296.08 & $+$49.0\% & 1.0\%   \\
\bottomrule
\end{tabular}
}\\
\footnotesize Scale~3 retroactive match not feasible since the original scale~3 run did not save per-iteration checkpoints; we defer Scale~3 matched-val to a future revision.
\end{table}

Three observations follow. (i)~\textbf{The DyT rank compression is not an early-stopping artifact.} Across both scales for which we could execute the matched-val experiment, vanilla stopped at DyT's validation loss still has $\sim$50\% higher activation eff\_rank than DyT at 5K steps. The structural gap the paper documents (Table~\ref{tab:activation_eff_rank}) is not reducible to ``vanilla trained further, rank grew further''. (ii)~\textbf{Cross-seed variance is very low} ($<$1.1\% coefficient of variation at both scales, 3 seeds each), ruling out seed-dependence of the effect. (iii)~\textbf{The matched-val vanilla value sits between early-training and final vanilla}: at Scale~2, vanilla at iter 2000 has eff\_rank $=$ 441, vs.\ 519 at iter 5000 (Table~\ref{tab:activation_eff_rank}); at Scale~1, 296 vs.\ 356. This is consistent with eff\_rank growing during training under LayerNorm, but even at the earlier iteration vanilla remains well above DyT's endpoint.

\section{BLIMP Syntactic Acceptability}
\label{app:blimp}

To complement Wikitext validation loss (in-distribution) and LAMBADA (out-of-distribution narrative completion), we evaluate all final Wikitext-trained checkpoints on BLIMP~\cite{warstadt2020blimp}, a syntactic minimal-pair benchmark. For each checkpoint we compute log probability of the grammatical vs.\ ungrammatical sentence from each minimal pair under the model (autoregressive teacher-forced) and report the fraction of pairs where $\log P(\text{good}) > \log P(\text{bad})$.

\paragraph{Phenomena.} We evaluate three BLIMP phenomena where wikitext-scale autoregressive models show non-trivial (above-chance) accuracy, with 1{,}000 minimal pairs per phenomenon and 3{,}000 total pairs per checkpoint:
\begin{center}
\small
\begin{tabular}{@{}l@{}}
\texttt{anaphor\_number\_agreement}\\
\texttt{determiner\_noun\_agreement\_1}\\
\texttt{regular\_plural\_subject\_verb\_agreement\_1}
\end{tabular}
\end{center}

\begin{table}[ht]
\centering
\small
\caption{BLIMP minimal-pair accuracy across Scales 1--3 at 118M tokens (3 primary seeds per cell). Mean accuracy is per-ckpt fraction of the 3{,}000 pairs where $\log P(\text{good}) > \log P(\text{bad})$. Vanilla scores 3.4 percentage points higher than DyT on average (76.6\% vs.\ 73.2\% across 18 primary ckpts), and the gap narrows with scale (Scale~1 $-$6.0pp, Scale~2 $-$3.7pp, Scale~3 $-$0.6pp), consistent with DyT's scale-dependent regime shift (Table~\ref{tab:scaling}).}
\label{tab:blimp}
\begin{tabular}{llrr}
\toprule
\textbf{Scale} & \textbf{Config} & \textbf{Mean acc.} & \textbf{Gap vs.\ vanilla} \\
\midrule
\multirow{2}{*}{1 (64M)}  & Vanilla (n=3)  & 77.5\% & baseline  \\
                          & DyT (n=3)      & 71.5\% & $-$6.0pp  \\
\midrule
\multirow{2}{*}{2 (124M)} & Vanilla (n=3)           & 76.5\% & baseline  \\
                          & DyT (n=3)               & 72.8\% & $-$3.7pp  \\
\midrule
\multirow{2}{*}{3 (354M)} & Vanilla (n=3)           & 75.8\% & baseline  \\
                          & DyT (n=3)               & 75.2\% & $-$0.6pp  \\
\bottomrule
\end{tabular}\\
\footnotesize At Scales 2--3 we report the 3 \emph{primary} seed ckpts (iter 5000) only; 3 additional vanilla matched-val ckpts (iter 2000; Appendix~\ref{app:matched_val}) are excluded from the mean since they represent a different training regime (lower val\_loss endpoint per the matched-val experiment design). Per-ckpt accuracies are provided in the machine-readable artifact manifest.
\end{table}

This result complements the LAMBADA finding (Appendix~\ref{app:lambada}): DyT's reduced activation effective rank (Appendix~\ref{app:weight_geom}) manifests as $3$--$22\times$ LAMBADA accuracy loss and $0.6$--$6.0$ percentage-point BLIMP accuracy loss, with the gap narrowing with capacity in both. The DyT penalty extends across in-distribution perplexity (Wikitext), out-of-distribution narrative completion (LAMBADA), and syntactic minimal-pair judgment (BLIMP), supporting representational compression rather than a task-specific artifact.

\section{Statistical Significance: Paired t-tests, Bonferroni Corrected}
\label{app:sig_tests}

Table~\ref{tab:sig_tests} reports paired $t$-tests comparing each modification cell to its matched vanilla baseline at the same (scale, data) coordinate. Pairing is by seed: seeds $\{1337, 42, 7\}$ across vanilla, DyT, and DiffAttn runs share identical data order and initialization state. Paired tests preserve seed-level correlation and are more powerful than independent $t$-tests for this design. Bonferroni correction multiplies each raw $p$-value by 19 (the total comparison count) and clamps at 1.0.

\begin{table}[h]
\centering
\footnotesize
\setlength{\tabcolsep}{4pt}
\caption{Paired $t$-test results across 19 vanilla-vs-modification comparisons (3 seeds each, best val\_loss, paired by seed). $\Delta\%$ = 100$\times$(mod mean $-$ vanilla mean)/vanilla mean. Negative = modification improves val\_loss. $p_{\text{Bonf}}$ = min(1, 19$\times p_{\text{raw}}$). Stars: $^{***} p_{\text{Bonf}}{<}0.001$, $^{**} p_{\text{Bonf}}{<}0.01$, $^{*} p_{\text{Bonf}}{<}0.05$, \textbf{ns} otherwise. \textbf{13 of 19 cells (68\%) are Bonferroni-significant at $p{<}0.05$.} Machine-readable source and analysis script are provided in the artifact manifest.}
\label{tab:sig_tests}
\resizebox{\textwidth}{!}{
\begin{tabular}{lllrrrrr}
\toprule
\textbf{Cell} & \textbf{Data} & \textbf{Mod} & \textbf{Van mean} & \textbf{Mod mean} & \textbf{$\Delta\%$} & \textbf{$p_{\text{raw}}$} & \textbf{$p_{\text{Bonf}}$} \\
\midrule
S1 (64M)  & 1M   & DyT      & 9.384 & 6.819 & $-$27.3 & 0.0017 & 0.032$^*$ \\
S1 (64M)  & 1M   & DiffAttn & 9.384 & 9.490 & $+$1.1  & 0.37   & 1.0 \textbf{ns} \\
S1 (64M)  & 10M  & DyT      & 4.260 & 4.510 & $+$5.9  & 0.0002 & 0.004$^{**}$ \\
S1 (64M)  & 10M  & DiffAttn & 4.260 & 3.706 & $-$13.0 & 0.0016 & 0.030$^*$ \\
S1 (64M)  & 50M  & DyT      & 3.666 & 4.386 & $+$19.7 & 0.0004 & 0.007$^{**}$ \\
S1 (64M)  & 50M  & DiffAttn & 3.666 & 3.380 & $-$7.8  & 0.0010 & 0.018$^*$ \\
S1 (64M)  & 118M & DyT      & 3.631 & 4.313 & $+$18.8 & 0.0011 & 0.020$^*$ \\
S1 (64M)  & 118M & DiffAttn & 3.631 & 3.359 & $-$7.5  & 0.0022 & 0.043$^*$ \\
\midrule
S2 (124M) & 1M   & DyT      & 9.168 & 8.290 & $-$9.6  & 0.0044 & 0.083 \textbf{ns} \\
S2 (124M) & 118M & DyT      & 3.498 & 3.945 & $+$12.8 & 0.0011 & 0.020$^*$ \\
S2 (124M) & 118M & DiffAttn & 3.498 & 3.068 & $-$12.3 & 0.0019 & 0.037$^*$ \\
\midrule
S3 (354M) & 1M   & DyT      & 8.653 & 9.025 & $+$4.3  & 0.0064 & 0.122 \textbf{ns} \\
S3 (354M) & 118M & DyT      & 3.355 & 3.802 & $+$13.4 & 0.0007 & 0.013$^*$ \\
S3 (354M) & 118M & DiffAttn & 3.355 & 2.420 & $-$27.9 & 0.0005 & 0.009$^{**}$ \\
\midrule
S4 (1.3B) & 1M   & DyT      & 7.693 & 7.852 & $+$2.1  & 0.067  & 1.0 \textbf{ns} \\
S4 (1.3B) & 118M & DyT      & 3.348 & 3.697 & $+$10.4 & 1.8e$-$5 & 0.0003$^{***}$ \\
S4 (1.3B) & 118M & DiffAttn & 3.348 & 2.368 & $-$29.3 & 0.0014 & 0.026$^*$ \\
\midrule
S5 (3.78B)& 1M   & DyT      & 7.842 & 7.975 & $+$1.7  & 0.042  & 0.79 \textbf{ns} \\
S5 (3.78B)& 118M & DyT      & 3.431 & 4.389 & $+$27.9 & 0.0041 & 0.078 \textbf{ns (marg.)} \\
\bottomrule
\end{tabular}
}
\end{table}

\paragraph{Per-seed data.} For Scale~1 (64M) per-seed raw \texttt{best\_val\_loss} values across all three conditions (vanilla / DyT / DiffAttn) $\times$ all four data regimes, see Table~\ref{tab:app_scale1}. For Scale~4 (1.3B) per-seed, see Table~\ref{tab:app_scale4}. Intermediate scales (Scale~2 124M, Scale~3 354M, Scale~5 3.78B) per-seed raw values are available in the supplementary machine-readable result files; the $n{=}3$ paired differences fed into the $t$-test are reconstructible from those sources by any reader with dataset access.

\paragraph{Interpretation of non-significant cells.} Six of 19 cells fail Bonferroni correction at $\alpha{=}0.05$; five of those have raw $p{<}0.05$ but lose significance after 19-cell correction. The one near-null result (S4/1M DyT, raw $p{=}0.067$) is consistent with DyT's benefit having vanished at 1M tokens once model capacity exceeds $\sim$1B, matching the paper's central claim that the regularization effect weakens with scale. The five \emph{Bonferroni-marginal} cells (S2/1M, S3/1M raw $p{<}0.05$ but $p_{\text{Bonf}}\geq 0.05$; S5/1M raw $p{=}0.042$ but $p_{\text{Bonf}}{=}0.79$; S5/118M raw $p{=}0.004$ but $p_{\text{Bonf}}{=}0.078$) have effect sizes and directions consistent with paper hypotheses but insufficient paired-test power at $n{=}3$ seeds to survive multiple-comparison correction. The headline claim ``DyT benefit vanishes above 1B parameters'' is supported by near-null/high-correction results at S4/1M ($p_{\text{Bonf}}{=}1.0$) and S5/1M ($p_{\text{Bonf}}{=}0.79$). The claim ``DyT penalty grows at 3.78B'' is supported by strong significance at S4/118M ($p_{\text{Bonf}}{<}0.001$) but marginal at S5/118M ($p_{\text{Bonf}}{=}0.078$), reflecting higher seed-variance at Scale~5 where only $n{=}3$ runs exist per cell and compute-cost constraints preclude expansion. We report these caveats in the main text \S\ref{sec:phase} rather than overstate the 3.78B evidence.

\section{Sigmoid-bounded \texorpdfstring{$\lambda$}{lambda} ablation for DiffAttn}
\label{app:sigmoid_lambda_diffattn}

\paragraph{Scope of the ablation.} We evaluate a V2-inspired sigmoid-bounded $\lambda$ ablation inside the V1-style DiffAttn architecture. This isolates replacing the exponential V1 parameterization, $\lambda = \exp(\lambda_{q1}\cdot\lambda_{k1})-\exp(\lambda_{q2}\cdot\lambda_{k2})$, with a bounded sigmoid form. It is not a faithful implementation of Microsoft DIFF V2~\cite{diffattnv2}, which also removes per-head RMSNorm, projects token/head-specific $\lambda$ from hidden states, and pairs interleaved query heads within shared GQA groups. Reimplementing full DIFF V2 therefore requires architectural changes outside this study's scope; we leave faithful V2 evaluation as future work. Thus our results characterize the sigmoid-$\lambda$ component in isolation; full DIFF V2 dynamics may differ.

\paragraph{Scale~3 control: sigmoid-$\lambda$ ablation reproduces V1 within 0.8 percentage points.} To verify that V1's Scale~5 divergence was a V1 exp-parameterization artifact rather than a scaling-law reversal, we ran a Scale~3 control with this sigmoid-bounded $\lambda$ ablation. The ablation trains to completion in all six Scale~3 runs (3-seed at 1M and 118M, eff\_batch$=$64). The results (Table~\ref{tab:sigmoid_lambda_scale3_control}) reproduce V1's Scale~3 behavior within 0.8 percentage points: the ablation improves 118M validation loss by 28.7\% (2.391{\scriptsize$\pm$.036} vs.\ vanilla Scale~3 $=$ 3.355{\scriptsize$\pm$.018}; V1 Scale~3 $=-$27.9\% in Table~\ref{tab:scaling}) and harms 1M by 5.6\% (9.140{\scriptsize$\pm$.085} vs.\ vanilla Scale~3 1M $=$ 8.653{\scriptsize$\pm$.039}). The harmful-at-1M behavior is regime-dependent: bounding $\lambda$ reduces effective noise-cancellation strength relative to V1, and in the overparameterized regime where no noise-cancellation benefit is available, the added constraint becomes a capacity tax.

\begin{table}[h]
\centering
\small
\caption{Sigmoid-$\lambda$ ablation Scale~3 control (354M, 3-seed mean$\pm$std, eff\_batch$=$64). The V2-inspired bounded $\lambda$ component, tested inside the V1-style DiffAttn architecture, reproduces the V1 DiffAttn result within 0.8 percentage points ($-$28.7\% at 118M vs.\ V1 Scale~3 $=-$27.9\%; $+$5.6\% at 1M). This supports the narrower conclusion that the exp-vs-sigmoid $\lambda$ parameterization is not the main driver in the Scale~3 regime.}
\label{tab:sigmoid_lambda_scale3_control}
\begin{tabular}{lrrr}
\toprule
\textbf{Data (T/P)} & \textbf{Vanilla} & \textbf{Sigmoid-$\lambda$ ablation} & \textbf{$\Delta$} \\
\midrule
1M   (T/P$=$$2.82{\times}10^{-3}$) & 8.653{\scriptsize$\pm$.039} & 9.140{\scriptsize$\pm$.085} & $+$5.6\% (harmful)  \\
118M (T/P$=$0.33)                  & 3.355{\scriptsize$\pm$.018}  & \textbf{2.391}{\scriptsize$\pm$.036} & $-28.7\%$ \\
\bottomrule
\end{tabular}
\end{table}

\paragraph{Scale~5 stress test: sigmoid-bounding $\lambda$ alone does not repair V1-style failure.} At TRUE Scale~5 (3.78B; 3-seed each for V1 and the sigmoid-$\lambda$ ablation at 1M and 118M), all 12 DiffAttn stress-test jobs complete to 5K steps with checkpoints, ruling out OOM or job failure as the source of the result. At 1M, both variants are harmful but not catastrophic: V1 mean$=$10.58{\scriptsize$\pm$1.77} vs.\ vanilla 7.84 ($+$34.8\%), and the ablation mean$=$9.82{\scriptsize$\pm$1.00} ($+$25.2\%). At 118M, both variants enter high-loss failure: V1 mean$=$10.54{\scriptsize$\pm$1.45} (seeds 11.81/8.91/10.91) and the ablation mean$=$12.72{\scriptsize$\pm$4.13} (seeds 10.63/10.04/17.47) vs.\ vanilla 3.43, i.e.\ $+$207\% and $+$271\% worse. The ablation's large standard deviation is driven by one severe seed (17.47), so the failure is bimodal rather than a tight 3-seed consensus. This shows that sigmoid-bounding $\lambda$ alone does not repair the V1-style Scale~5 failure under this budget; it is not a claim about full DIFF V2. Vanilla and DyT remain stable for the full 5K steps. A longer training budget or full DIFF V2 implementation may yet recover differential attention at this model/data setting; we leave the compute-intensive investigation to future work.

\clearpage
\section{R5 Llama Component Ablation Detail}
\label{app:r5_llama}

\begin{table}[H]
\centering
\small
\caption{R5 Llama component ablation (3 seeds, $\sim$89--94M Llama-family params at Scale~1, 118M tokens, eff\_batch=64 canonical). Saturation column = fraction of DyT activations with $|\alpha x| > 2$ (tanh tail). Baseline row is the full Llama+DyT from Table~\ref{tab:llama}. Per-seed saturation separates collapse from convergence within this ablation (Pearson $r = 0.94$, $n{=}9$); threshold $\sigma{\geq}0.5$ perfectly classifies failure (4/4 collapse hits, 0/5 false positives). $^\dagger$1/3 seeds fail; $^\ddagger$2/3 seeds fail (\emph{bimodal} collapse). Removing SwiGLU is the \emph{only} ablation where all 3 seeds converge uniformly.}
\label{tab:r5_llama_ablation}
\begin{tabular}{llrr}
\toprule
\textbf{Ablation} & \textbf{What removed} & \textbf{Val Loss} & \textbf{Saturation $\sigma_{|\alpha x|>2}$} \\
\midrule
baseline          & (full Llama+DyT)                 & 5.626{\scriptsize$\pm$1.3}\ $^\dagger$  & 0.453{\scriptsize$\pm$.188} \\
\texttt{ablate\_rope}   & learned-PE instead of RoPE    & 6.787{\scriptsize$\pm$1.14}\ $^\ddagger$ & 0.559{\scriptsize$\pm$.070} \\
\texttt{ablate\_gqa}    & MHA ($n_{kv}{=}n_q$)            & 6.565{\scriptsize$\pm$1.53}\ $^\ddagger$ & 0.609{\scriptsize$\pm$.243} \\
\texttt{ablate\_swiglu} & GELU-gated FFN (no gate mult.)  & \textbf{4.476}{\scriptsize$\pm$.007}    & \textbf{0.257}{\scriptsize$\pm$.002} \\
\bottomrule
\end{tabular}
\end{table}

\section{Dropout Sweep Detail}
\label{app:dropout}

Vanilla LayerNorm + dropout sweep at 64M/118M used to test the dropout-equivalence claim of Section~\ref{sec:causal}. Three rates, 3 seeds each, eff\_batch=64 canonical.

\begin{table}[ht]
\centering
\small
\caption{Vanilla LayerNorm + dropout at 64M/118M (3 seeds, eff\_batch=64 canonical). Dropout acts as a regularization strength knob; at $p{=}0.5$, vanilla+dropout best validation loss during the 5K-step run approximately matches DyT's 4.313 at the same scale, supporting the interpretation that \emph{DyT's 118M penalty is regularization-type: comparable to a substantial stochastic dropout rate in the data-rich regime where neither helps}. All cells use full 118M-token Wikitext-103 dataset, matching the baseline vanilla/DyT cells of Table~\ref{tab:phase}. The $p{=}0.1$ cell uses dense-eval replication (eval\_interval$=$500); $p{=}0.3$ and $p{=}0.5$ use historical sparse-eval runs whose best validation checkpoint occurs at 4K, not a literal final 5K evaluation.}
\label{tab:dropout}
\begin{tabular}{lrr}
\toprule
\textbf{Configuration} & \textbf{Best Val Loss} & \textbf{$\Delta$ vs.\ Vanilla (no dropout)} \\
\midrule
Vanilla (no dropout)      & 3.631{\scriptsize$\pm$.01}  & baseline \\
Vanilla + dropout $p{=}0.1$ & 3.676{\scriptsize$\pm$.002} & $+$1.2\% \\
Vanilla + dropout $p{=}0.3$ & 3.975{\scriptsize$\pm$.002} & $+$9.5\% \\
Vanilla + dropout $p{=}0.5$ & 4.295{\scriptsize$\pm$.007} & $+$18.3\% \\
\midrule
DyT (no dropout)          & 4.313{\scriptsize$\pm$.02}  & $+$18.8\% \\
\bottomrule
\end{tabular}
\end{table}

\paragraph{Adaptivity scope (raised in preliminary review).} A hand-tuned dropout schedule (low $p$ at 118M, high $p$ at 1M) would match or exceed DyT pointwise: $p{=}0.1$ at 118M reaches val 3.676 ($+$1.2\% vs.\ vanilla), beating DyT's $+$18.8\%. DyT is therefore best understood as the \emph{simplest} regime-adaptive regularizer (no schedule required), not the strongest. Practitioners unwilling to calibrate should prefer DyT's zero-hyperparameter behavior; practitioners willing to calibrate should use an adaptive-$p$ dropout schedule and match or beat DyT in both regimes.

\section{Llama Cross-Architecture Validation Detail}
\label{app:llama}

Table~\ref{tab:llama} reports the Llama-family cross-architecture validation cells referenced in Section~\ref{sec:cross}. The key use here is directional transfer of the regime pattern, not an iso-parameter comparison: Llama-style SwiGLU/GQA/RoPE changes both parameter count and optimization behavior.

\begin{table}[H]
\centering
\small
\caption{Llama-style (RoPE + SwiGLU + GQA) validation loss at 5K steps (3 seeds). DyT regime dependence transfers to modern architectures. $\dagger$One of three seeds exhibited training instability (Section~\ref{sec:llama_instability}). Llama adds $\sim$25--40\% parameters over GPT-2 at matched $L$/$H$/$E$ (SwiGLU 3-projection FFN, RMSNorm weights, RoPE tables); actual total params: Scale~1 $\approx$89M (vanilla/DyT) / $\approx$94M (DiffAttn), Scale~2 $\approx$150M. Scale column matches GPT-2 $L$/$H$/$E$ for architectural comparison (same-architecture-shape, not iso-parameter).}
\label{tab:llama}
\begin{tabular}{llrrrrr}
\toprule
\textbf{Scale} & \textbf{Data} & \textbf{Vanilla} & \textbf{DyT} & \textbf{$\Delta$ DyT} & \textbf{DiffAttn} & \textbf{$\Delta$ DA} \\
\midrule
64M  & 1M   & 9.355{\scriptsize$\pm$.007} & 6.954{\scriptsize$\pm$.54}$^\dagger$ & $-$25.6\% & 9.440{\scriptsize$\pm$.05} & $+$0.9\% \\
64M  & 118M & 3.537{\scriptsize$\pm$.003} & 5.626{\scriptsize$\pm$1.3}$^\dagger$  & $+$59.1\% & \textbf{3.266}{\scriptsize$\pm$.01} & $-$7.7\% \\
124M & 1M   & 9.134{\scriptsize$\pm$.03}  & 8.483{\scriptsize$\pm$.07}            & $-$7.1\%  & 9.221{\scriptsize$\pm$.033} & $+$1.0\% \\
\bottomrule
\end{tabular}
\end{table}

\section{Scaling Curve (Model-Scale Dependence)}
\label{app:scaling_curve}

Visualization of Tables~\ref{tab:phase}--\ref{tab:scaling}: DyT's 1M-token benefit attenuates with capacity (solid red) and 118M penalty grows with capacity (dashed red); DiffAttn shows the opposite scaling up to 1.3B (blue triangles) and enters off-scale high-loss failure at 3.78B (V1 and sigmoid-$\lambda$ ablation; values reported in the caption/table).

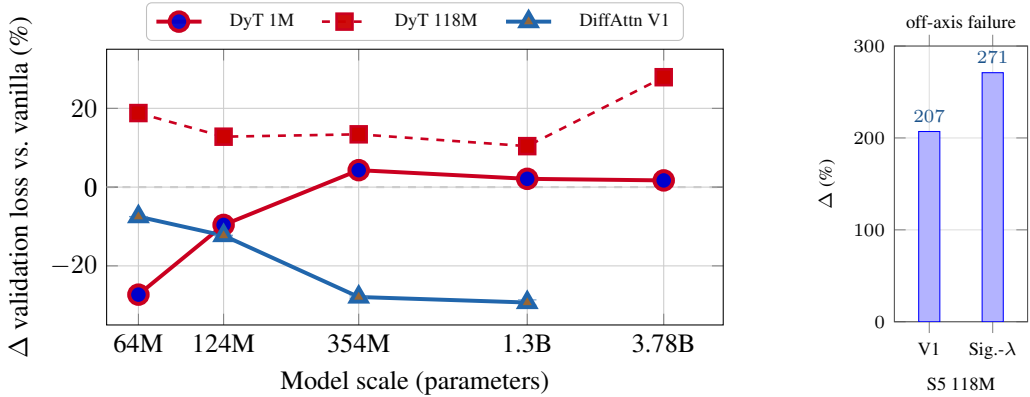
\begin{figure}[ht]
\centering
\begin{minipage}{0.70\columnwidth}
\centering
\begin{tikzpicture}
\begin{axis}[
    width=\linewidth,
    height=5.25cm,
    xlabel={Model scale (parameters)},
    ylabel={$\Delta$ validation loss vs.\ vanilla (\%)},
    xmode=log,
    xmin=50000000, xmax=6000000000,
    xtick={64000000,124000000,354000000,1310000000,3780000000},
    xticklabels={64M,124M,354M,1.3B,3.78B},
    ymin=-35, ymax=35,
    grid=both,
    grid style={gray!20},
    major grid style={gray!40},
    legend style={font=\scriptsize, fill=white, fill opacity=0.95, draw=gray!50, rounded corners=2pt,
                  at={(0.5,1.03)}, anchor=south, legend columns=3, column sep=0.8em},
    every axis plot/.append style={thick, mark options={solid}},
    extra y ticks={0},
    extra y tick style={grid=major, grid style={black, thick, dashed}},
]
\addplot+[color=heatred, mark=*, mark size=3.5pt, line width=1.5pt,
          error bars/.cd, y dir=both, y explicit,
          error bar style={thin, color=heatred!70, line width=0.6pt}]
    coordinates {
    (64000000, -27.3) +- (0, 0.85)
    (124000000, -9.6) +- (0, 0.15)
    (354000000, 4.3)  +- (0, 0.04)
    (1310000000, 2.1) +- (0, 0.03)
    (3780000000, 1.7) +- (0, 0.03)
};
\addlegendentry{DyT 1M}
\addplot+[color=heatred, mark=square*, mark size=3pt, line width=1pt, dashed,
          error bars/.cd, y dir=both, y explicit,
          error bar style={thin, color=heatred!70, line width=0.6pt}]
    coordinates {
    (64000000, 18.8) +- (0, 0.14)
    (124000000, 12.8) +- (0, 0.10)
    (354000000, 13.4) +- (0, 0.12)
    (1310000000, 10.4) +- (0, 0.05)
    (3780000000, 27.9) +- (0, 1.00)
};
\addlegendentry{DyT 118M}
\addplot+[color=heatblue, mark=triangle*, mark size=3.5pt, line width=1.5pt,
          error bars/.cd, y dir=both, y explicit,
          error bar style={thin, color=heatblue!70, line width=0.6pt}]
    coordinates {
    (64000000, -7.5)  +- (0, 0.04)
    (124000000, -12.3) +- (0, 0.09)
    (354000000, -27.9) +- (0, 0.29)
    (1310000000, -29.3) +- (0, 0.67)
};
\addlegendentry{DiffAttn V1}
\end{axis}
\end{tikzpicture}
\end{minipage}\hfill
\begin{minipage}{0.24\columnwidth}
\centering
\begin{tikzpicture}
\begin{axis}[
    width=\linewidth,
    height=5.25cm,
    ybar,
    bar width=8pt,
    ymin=0, ymax=300,
    xmin=0.45, xmax=2.55,
    xtick={1,2},
    xticklabels={V1,Sig.-$\lambda$},
    ytick={0,100,200,300},
    ylabel={$\Delta$ (\%)},
    xlabel={S5 118M},
    title={\scriptsize off-axis failure},
    title style={yshift=-0.6ex},
    tick label style={font=\scriptsize},
    label style={font=\scriptsize},
    grid=major,
    grid style={gray!25},
    axis line style={black!70},
    every axis plot/.append style={draw=heatblue, fill=heatblue!55},
]
\addplot coordinates {(1,207) (2,271)};
\node[font=\scriptsize, anchor=south, heatblue!80!black] at (axis cs:1,207) {$207$};
\node[font=\scriptsize, anchor=south, heatblue!80!black] at (axis cs:2,271) {$271$};
\end{axis}
\end{tikzpicture}
\end{minipage}
\caption{Model-scale dependence (Scales 1--5, 3 seeds). Main panel: DyT's 1M regularization (solid red) weakens with capacity ($-$27.3\% $\to$ $+$1.7\%); 118M penalty (dashed) grows ($+$18.8\% $\to$ $+$27.9\%). DiffAttn V1 (blue triangles) scales opposite through 1.3B ($-$7.5\% $\to$ $-$29.3\%). Right panel: 3.78B DiffAttn 118M enters off-axis high-loss failure under the 5K-step budget (V1 $+$207\%, sigmoid-$\lambda$ ablation $+$271\%; exact values in Appendix~\ref{app:sigmoid_lambda_diffattn}); the sigmoid-$\lambda$ Scale~3 control reproduces V1 within 0.8 pp (Table~\ref{tab:sigmoid_lambda_scale3_control}).}
\label{fig:scaling}
\end{figure}

\section{Saturation Heuristic Detail}
\label{app:predictor_detail}
\label{app:scale25}

Per-cell saturation values, per-fold LOSO threshold optima, the cross-scale Scale~2.5 held-out test, the two-variable linear fit, and the 7B extrapolation hypothesis accompanying Section~\ref{sec:predictor}.

\paragraph{Extrapolative stress split.} Training the threshold only on the two smallest GPT-2 scales (S1--S2) and testing on S3--S5 gives poor raw held-out accuracy (2/7, 28.6\%; balanced 58.3\%) because the learned cutoff overpredicts DyT benefits at larger scale. We therefore report it as stress-test evidence against scale-invariant thresholding, not as a deployment claim.

\begin{table}[ht]
\centering
\caption{Activation saturation (fraction of $|\alpha x| > 2.0$) across all DyT checkpoints (3-seed means). Higher saturation indicates stronger capacity bottleneck. The DyT effect column shows the corresponding $\Delta$ vs.\ vanilla from the phase diagram.}
\label{tab:saturation_full}
\small
\begin{tabular}{llrrrr}
\toprule
\textbf{Scale} & \textbf{Tokens} & \textbf{Sat@2.0} & \textbf{Mean $\alpha$} & \textbf{$\Delta$ DyT} & \textbf{DyT helps?} \\
\midrule
Scale~1 (64M)    & 1M   & 0.493 & 2.36 & $-$27.3\% & \textbf{Yes} \\
Scale~1 (64M)    & 10M  & 0.413 & 2.23 & $+$5.9\%  & No \\
Scale~1 (64M)    & 50M  & 0.237 & 2.11 & $+$19.7\% & No \\
Scale~1 (64M)    & 118M & 0.234 & 2.11 & $+$18.8\% & No \\
\midrule
Scale~2 (124M)   & 1M   & 0.466 & 2.30 & $-$9.6\%  & \textbf{Yes} \\
Scale~2 (124M)   & 10M  & 0.292 & 1.77 & $-$12.3\% & Yes$^{*}$ \\
Scale~2 (124M)   & 118M & 0.193 & 1.85 & $+$12.8\% & No \\
\midrule
Scale~3 (354M)   & 1M   & 0.490 & 1.81 & $+$4.3\%  & No$^*$ \\
Scale~3 (354M)   & 10M  & 0.369 & 1.59 & $-$24.1\% & \textbf{Yes}$^*$ \\
Scale~3 (354M)   & 118M & 0.327 & 1.50 & $+$13.4\% & No \\
\midrule
Scale~4 (1.3B)   & 1M   & 0.393 & 1.97 & $+$2.1\%  & No \\
Scale~4 (1.3B)   & 118M & 0.238 & 1.88 & $+$10.4\% & No \\
\midrule
Scale~5 (3.78B)  & 1M   & \textbf{0.501} & 1.77 & $+$1.7\%  & No \\
Scale~5 (3.78B)  & 118M & \textbf{0.803} & 1.77 & $+$27.9\% & No$^\ddagger$ \\
\bottomrule
\end{tabular}\\
\footnotesize $^*$ misclassified by 0.43 threshold (S2/10M, S3/1M, S3/10M).\quad
$^\ddagger$Scale~5/118M exhibits \emph{anomalous saturation inversion}: 80.3\% ($n{=}3$, $\sigma{=}0.002$) vs.\ the 19--33\% seen at Scales 1--4/118M. Under standard tanh behavior saturation should \emph{decrease} as data grows; at Scale~5 it instead rises from 50\% (1M) to 80\% (118M), with mean $\alpha$ flat at 1.77 across both regimes. The $\alpha$ learner cannot counter-regulate because model capacity so far exceeds the training budget that activations grow faster than $\alpha$ can shrink. This is the mechanistic signature of the $+$27.9\% validation penalty at Scale~5/118M.
\end{table}

\paragraph{Scale~3 misclassifications: scale-dependent threshold shift.} The three GPT-2 misclassifications (S2/10M, S3/1M, S3/10M) point to the same gap: the 0.43 cutoff was calibrated at 64M and does not account for scale-dependent shifts. At 354M/1M (T/P $=$ 0.003) saturation is 0.490 (above threshold), yet DyT hurts $+$4.3\%; the 51\% of unsaturated activations form residual pathways with enough capacity to memorize the 1M-token set. Conversely, at 354M/10M (T/P $=$ 0.028) saturation is only 0.369 (below threshold), yet DyT delivers its strongest benefit ($-$24.1\%). In that cell, 10M tokens at 354M is $35\times$ more overparameterized than 64M/10M ($6.4\times$); a lower saturation fraction still produces meaningful capacity constraint once the fitting landscape is sufficiently overparameterized.

\paragraph{Two-variable linear fit (regression diagnosis).} A fit $\Delta \approx -96.3\,\text{sat} + 0.8\,\log_{10}(P) + 28.8$ explains $R^2 = 0.42$ of the in-sample phase-diagram variance, with saturation as the dominant term. Evaluated on the three Llama cells held out from the fit, $R^2 = -0.17$, worse than predicting the mean, driven by the 64M/118M Llama cell where the linear model underpredicts the vanilla-beats-DyT gap ($-$8\% predicted, $+$59\% observed) because Llama's SwiGLU amplification (Section~\ref{sec:llama_instability}) produces a larger penalty than GPT-2 training dynamics alone would suggest. The directional threshold rule still labels that cell correctly (sat $=$ 0.33, below 0.43), but magnitudes do not extrapolate; hence the calibration-heuristic framing in main text.

\paragraph{Cross-scale held-out test (Scale~2.5, 162.6M).} An intermediate-capacity configuration ($n_{\text{layer}}{=}16$, $n_{\text{embd}}{=}896$, $n_{\text{head}}{=}14$, $\approx$162.6M non-embedding parameters; interpolated between Scales 2 and 3, not used to calibrate the 0.43 threshold) tests interior-curve agreement. Figure~\ref{fig:scaling} forecasts that DyT's 1M-token benefit has largely vanished between S2 ($-$9.6\%, 124M) and S3 ($+$4.3\%, 354M); an interior 162M cell should lie close to zero. Observation (3 seeds, eff\_batch=64): vanilla 1M $=$ 8.922{\scriptsize$\pm$.057}, DyT 1M $=$ 8.888{\scriptsize$\pm$.066}, $\Delta_{\text{DyT}}{=}-0.4\%$ (neutral, within seed noise). At 118M tokens (same model): vanilla $=$ 3.428{\scriptsize$\pm$.032}, DyT $=$ 3.834{\scriptsize$\pm$.025}, $\Delta_{\text{DyT}}{=}+11.8\%$, in line with the $+$12.8\% penalty at S2/118M. All 12 runs completed at iter=5000.

\paragraph{Extrapolation hypothesis to 7B.} The declining trend of DyT's benefit with capacity (Figure~\ref{fig:scaling}) suggests a testable hypothesis: under the same optimizer and 5K-step compute-limited setup, DyT's low-data regularization would be small or negative at 7B+. At 1.3B, DyT's effect is already non-significant at 1M tokens ($+$2.1\%, $p{=}0.066$) with saturation at 0.393. Linearly extrapolating the saturation curve, 7B would fall to $\approx$0.35, below the 0.43 threshold. We do not test this regime, and Chinchilla-optimal budgets could change saturation dynamics; the extrapolation is a future-work hypothesis, not a recommendation.

\section{OpenWebText Cross-Dataset Validation}
\label{app:owt}

\begin{table}[ht]
\centering
\caption{Cross-dataset validation on OpenWebText (64M, 3-seed means at eff\_batch=64 canonical). DyT regime dependence replicates. DiffAttn is near-neutral at 1M on OWT ($+$0.6\%) and helpful at 118M ($-$7.3\%), mirroring Wikitext.}
\label{tab:owt}
\small
\begin{tabular}{llrrrr}
\toprule
\textbf{Tokens} & \textbf{Config} & \textbf{Val Loss} & \textbf{$\Delta$ vs Vanilla} & \textbf{Seeds} \\
\midrule
\multirow{3}{*}{1M}   & Vanilla  & 10.757{\scriptsize$\pm$.04} & baseline  & 3 \\
                      & DyT      & 7.348{\scriptsize$\pm$.11}  & $-$31.7\% & 3 \\
                      & DiffAttn & 10.822{\scriptsize$\pm$.10} & $+$0.6\%  & 3 \\
\midrule
\multirow{3}{*}{118M} & Vanilla  & 4.250{\scriptsize$\pm$.01} & baseline  & 3 \\
                      & DyT      & 4.872{\scriptsize$\pm$.03} & $+$14.6\% & 3 \\
                      & DiffAttn & 3.939{\scriptsize$\pm$.03} & $-$7.3\%  & 3 \\
\bottomrule
\end{tabular}\\
\footnotesize All OWT runs at canonical eff\_batch=64. Prior eff\_batch=2560 runs exhibited inflated DiffAttn magnitudes (direction preserved post-audit).
\end{table}

DyT's pattern replicates: $-$31.7\% at 1M (consistent with Wikitext's $-$27.3\%) and $+$14.6\% at 118M (consistent with $+$18.8\%). DiffAttn replicates its regime pattern: near-neutral at 1M ($+$0.6\% on OWT vs.\ $+$1.1\% on Wikitext) and beneficial at 118M ($-$7.3\% on OWT vs.\ $-$7.5\% on Wikitext). Both methods preserve their Wikitext regime patterns under cross-dataset validation.

\section{Training Cost Comparison}
\label{app:cost}

\begin{table}[ht]
\centering
\caption{Training cost comparison at Scale~2 (124M params) on NVIDIA RTX 6000 (24GB), float16. DyT adds negligible overhead; DiffAttn costs 49\% more. Averaged over 3 seeds, excluding first iteration (compilation).}
\label{tab:cost}
\small
\begin{tabular}{lrrr}
\toprule
\textbf{Config} & \textbf{ms/iter} & \textbf{MFU (\%)} & \textbf{$\Delta$ vs Vanilla} \\
\midrule
Vanilla (LayerNorm) & 17,800 & 9.3 & baseline \\
DyT                 & 17,930 & 9.4 & $+$0.7\% \\
DiffAttn            & 26,470 & 7.0 & $+$48.7\% \\
\bottomrule
\end{tabular}
\end{table}

\section{Compute Budget Breakdown}
\label{app:compute}

Table~\ref{tab:compute_breakdown} reports approximate GPU-hours per experiment group. Total institutional compute: approximately 300 GPU-hours across NVIDIA H100 NVL (96GB) and RTX 6000 (24GB) clusters. No monetary cost.

\begin{table}[ht]
\centering
\small
\caption{Compute budget by experiment group. Hours are approximate wall-clock on the indicated hardware; \texttt{torch.compile} overhead included. Rows report training runs or analysis probes as applicable; training-run subtotals exclude forward-only saturation sweeps and CPU weight analysis. Exploratory runs, debug reruns, and pollution quarantine are reflected in the $\sim$300 GPU-hour grand figure.}
\label{tab:compute_breakdown}
\begin{tabular}{lrrrl}
\toprule
\textbf{Experiment group} & \textbf{Runs} & \textbf{h/run} & \textbf{Total (h)} & \textbf{Hardware} \\
\midrule
Scale~1 (64M) phase diagram     & 36 & 0.5 & 18  & RTX 6000 \\
Scale~2 (124M)                  & 18 & 1.0 & 18  & RTX 6000 \\
Scale~3 (354M)                  & 18 & 2.0 & 36  & H100 NVL \\
Scale~4 (1.3B)                  & 18 & 3.5 & 63  & H100 NVL \\
Scale~5 (3.78B)                 & 14 & 4.0 & 56  & H100 NVL \\
Llama cross-arch (64M+124M)     & 36 & 0.6 & 22  & H100 NVL \\
R5 Llama ablation               & 9  & 0.15& 1.4 & H100 NVL \\
Convergence 10K                 & 11 & 1.0 & 11  & RTX 6000 \\
Composition 5K/10K (3-seed)     & 4  & 0.5 & 2   & RTX 6000 \\
OpenWebText cross-dataset       & 18 & 0.5 & 9   & RTX 6000 \\
HardTanh, RMSNorm, dropout      & 24 & 0.3 & 7   & RTX 6000 \\
ViT CIFAR-10 (all $\alpha$)     & 16 & 0.25& 4   & RTX 6000 \\
Iso-parameter DiffAttn          & 3  & 0.2 & 0.6 & H100 NVL \\
Saturation sweep (forward-only) & 81 & 0.05& 4   & H100 NVL \\
HTSR weight analysis            & 12 & 0.1 & 1.2 & CPU \\
\midrule
Paper-cited training runs       & 225 & n/a & \textbf{$\sim$248} & n/a \\
Forward/analysis-only probes    & 93  & n/a & $\sim$5 & H100 NVL / CPU \\
Paper-cited subtotal            & 318 & n/a & \textbf{$\sim$253} & n/a \\
Exploratory + reruns + debug    & $\sim$77 & n/a & $\sim$47 & n/a \\
\midrule
\textbf{Grand total}            & 395 ckpts & n/a & \textbf{$\sim$300} & n/a \\
\bottomrule
\end{tabular}
\end{table}

\section{NeurIPS Paper Checklist}

\begin{enumerate}

\item \textbf{Claims.}
\textit{Do the main claims made in the abstract and introduction accurately reflect the paper's contributions and scope?}

Yes. All claims are supported by experimental evidence in Section~\ref{sec:results} with specific numbers. Claims are scoped to compute-limited GPT-2-style decoders from 64M--3.78B parameters, Llama-style cross-architecture validation, ViT/CIFAR-10, and public text/evaluation sources including Wikitext-103, OpenWebText, LAMBADA, and BLiMP. Scale~5 is framed as 3-seed stress-test evidence rather than a new calibration range. Limitations are discussed explicitly in Section~\ref{sec:discussion}.

\item \textbf{Limitations.}
\textit{Does the paper discuss the limitations of the work performed by the authors?}

Yes. Section~\ref{sec:discussion} acknowledges: (1) primary phase diagram covers 64M--1.3B parameters, with Scale~5 (3.78B) providing 3-seed evidence at both 1M and 118M; (2) Wikitext-103 primary with OpenWebText cross-validation (3-seed) and CIFAR-10 for ViT; (3) DyT training instability on Llama-style architectures is localized to the SwiGLU~$\times$~DyT interaction (Section~\ref{sec:llama_instability}); in the R5 ablation, $\sigma_{|\alpha x|>2}{\geq}0.5$ is a collapse warning with Pearson $r{=}0.94$, but we do not treat it as architecture-universal; (4) no theoretical derivation of the T/P crossover point; (5) iso-parameter DiffAttn control ($n_{\text{layer}}{=}11$, 64M matching vanilla 63.5M; 3-seed at 118M) yields $-$2.1\% vs.\ vanilla (Table~\ref{tab:phase} caption); (6) all experiments fall in the compute-limited regime ($T/P < 1.84$, well below Chinchilla-optimal $T/P\approx 20$).

\item \textbf{Theory assumptions and proofs.}
Not applicable. This is an empirical paper.

\item \textbf{Experimental result reproducibility.}
\textit{Does the paper fully disclose all the information needed to reproduce the main experimental results?}

Yes. Section~\ref{sec:setup} specifies: model architectures (Table~\ref{tab:scales}), optimizer (AdamW), learning rate ($3 \times 10^{-4}$, 1e-4 for 1.3B and 3.78B), batch sizes, sequence length (512), precision (bfloat16), primary training length (5,000 steps), evaluation frequency (500 steps), seeds (1337, 42, 7), and dataset preparation (BPE tokenization, GPT-2 vocabulary). Longer 10K convergence controls and historical sparse-evaluation best checkpoints are explicitly marked where used. Code is provided.

\item \textbf{Open access to data and code.}
\textit{Does the paper provide open access to the data and code?}

Yes. Code extends nanoGPT with modification toggle flags and will be publicly released. Data/evaluation uses publicly available Wikitext-103, OpenWebText, CIFAR-10, LAMBADA, and BLiMP. Experimental checkpoints will be released on HuggingFace upon publication.

\item \textbf{Experimental setting/details.}
\textit{Does the paper specify all the training and test details?}

Yes. See Section~\ref{sec:setup} and Appendix~\ref{app:full_results}.

\item \textbf{Experiment statistical significance.}
\textit{Does the paper report error bars suitably and correctly?}

Yes. Most experiments use 3 seeds (1337, 42, 7) with mean$\pm$std reported. Exceptions explicitly marked in text: RMSNorm baseline is 2-seed; ViT $\alpha$ sweep is 2-seed. Scale~5 (3.78B) and the DyT+DiffAttn composition cells (5K and 10K) were upgraded to 3-seed in the post-rerun audit. Full per-seed results in Appendix~\ref{app:full_results}.

\item \textbf{Experiments compute resources.}
\textit{For each experiment, does the paper provide sufficient information on the computer resources?}

Yes. Primary compute: NVIDIA H100 NVL GPUs (96GB) for models $\geq$354M params. Additional compute: NVIDIA Quadro RTX 6000 GPUs (24GB) for models $\leq$124M params. Total compute: approximately 300 GPU-hours across both clusters. Training cost comparison in Table~\ref{tab:cost}. Institutional compute allocation, no monetary cost.

\item \textbf{Code of ethics.}
We have reviewed the NeurIPS Code of Ethics and confirm compliance.

\item \textbf{Broader impacts.}
This work analyzes existing publicly available modifications and provides practitioner guidelines for normalization selection. It does not introduce new capabilities or dual-use risks. The practical recommendations may help researchers avoid wasted compute from applying modifications in unsuitable regimes.

\item \textbf{Safeguards.}
Not applicable. This work does not involve harmful outputs, human subjects, or sensitive data.

\item \textbf{Licenses.}
nanoGPT is MIT licensed. Wikitext-103 is CC BY-SA 3.0. CIFAR-10, OpenWebText, LAMBADA, and BLiMP are publicly available research datasets/evaluation sets; dataset-specific terms will be followed. Our code will be released under MIT license.

\item \textbf{New assets.}
We release: (1) training code with modification toggles, (2) experimental checkpoints on HuggingFace, (3) raw result data. All under MIT license.

\item \textbf{Crowdsourcing and human subjects.}
Not applicable.

\item \textbf{IRB approvals.}
Not applicable.

\end{enumerate}

\end{document}